\documentclass[10pt,twocolumn,letterpaper]{article}

\usepackage[pagenumbers]{cvpr} 

\usepackage{booktabs}
\usepackage[pagebackref,breaklinks,colorlinks,citecolor=cvprblue,hyperfootnotes=false]{hyperref}
\usepackage[utf8]{inputenc} 
\usepackage[T1]{fontenc}    
\usepackage{url}            
\usepackage{booktabs}       
\usepackage{amsfonts}       
\usepackage{amsmath}       
\usepackage{nicefrac}       
\usepackage{microtype}      
\usepackage[table]{xcolor}
\usepackage{multirow}
\usepackage{graphicx}
\usepackage{amssymb}
\usepackage{pifont}
\usepackage{tikz}
\usepackage{subcaption}
\usepackage{array}
\newcommand{\cmark}{\ding{51}}%
\newcommand{\xmark}{\ding{55}}%

\newcommand{\sm}[1]{\scriptsize{#1}}

\usepackage{makecell}
\definecolor{gold}{rgb}{1.0, 0.87, 0.0}
\definecolor{silver}{rgb}{0.75, 0.75, 0.75}
\definecolor{bronze}{rgb}{0.8, 0.5, 0.2}
\definecolor{ours_text_blue}{RGB}{204, 229, 255}
\definecolor{ours_green}{RGB}{204, 255, 204}

\newcommand{\f}[1]{\textbf{#1}}
\newcommand{\fc}[1]{\cellcolor{ours_text_blue}{#1}}
\newcommand{\scolor}[1]{\cellcolor{ours_green}{#1}}

\definecolor{cvprblue}{rgb}{0.21,0.49,0.74}

\def\paperID{69} 
\def\confName{3DV\xspace}
\def\confYear{2025\xspace}

\begin{document}
\twocolumn[{%
\renewcommand\twocolumn[1][]{#1}%
\title{MonoPatchNeRF: Improving Neural Radiance Fields with Patch-based Monocular Guidance} 




\author{
Yuqun Wu\textsuperscript{1*} \qquad \qquad
Jae Yong Lee\textsuperscript{1*} \qquad \qquad
Chuhang Zou\textsuperscript{2} \\
Shenlong Wang\textsuperscript{1} \qquad \qquad
Derek Hoiem\textsuperscript{1} \\
\textsuperscript{1}University of Illinois at Urbana-Champaign \\
\textsuperscript{2}Amazon Inc.
}


\maketitle

\begin{center}
    \centering
    \scriptsize
    \captionsetup{type=figure}
    \includegraphics[width=0.99\textwidth]{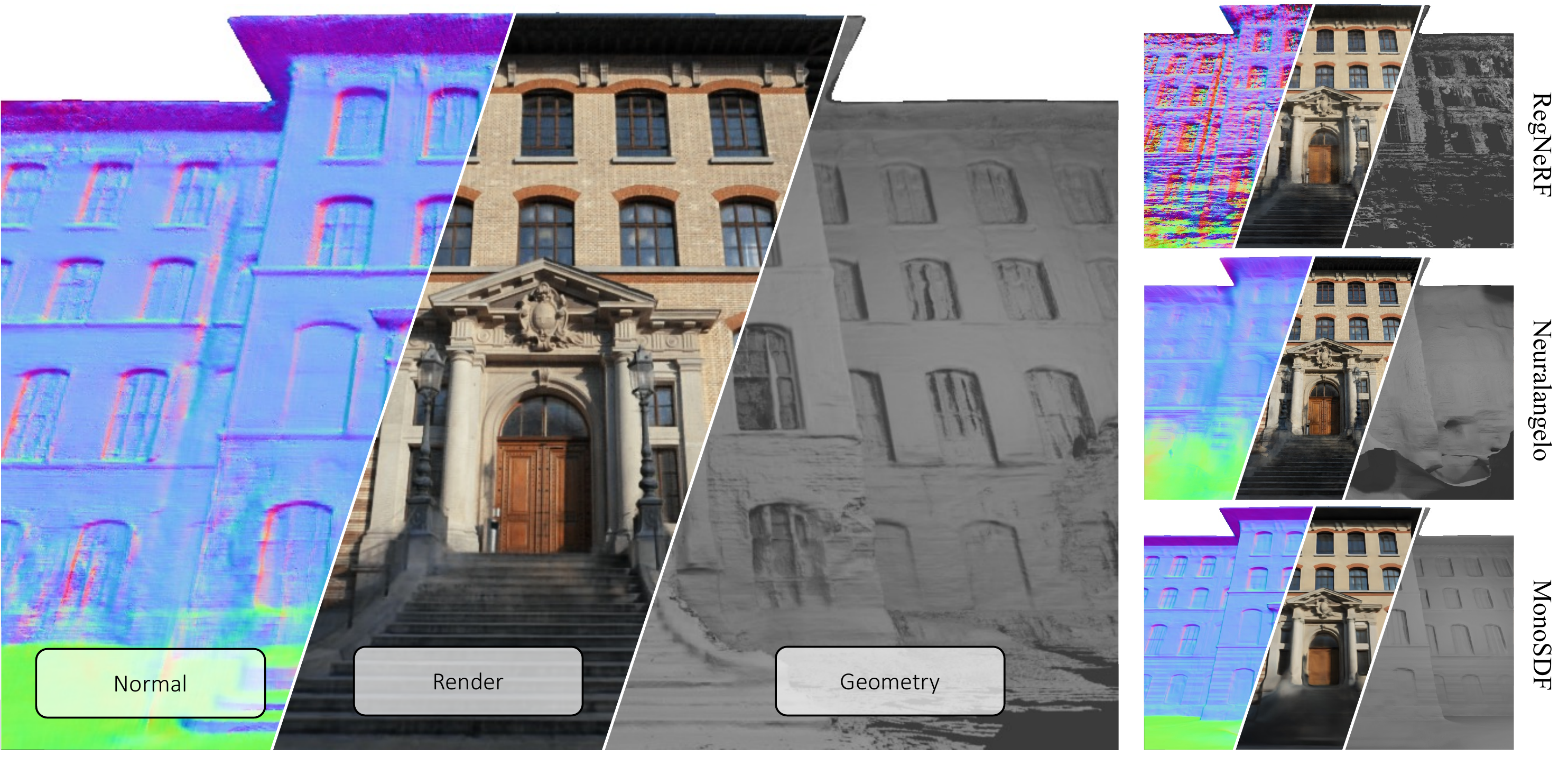}
    \caption{We present \textbf{MonoPatchNeRF}~(left) on a large-scale scene \textit{facade} that contains 68 input views. Our method renders realistic images and accurate normals from the test view and reconstructs the complete mesh compared to baselines~\cite{niemeyer2022regnerf,li2023neuralangelo,yu2022monosdf}.}
\label{fig:teaser}

        
\end{center}%
}]

\def\thefootnote{$\ast$}\footnotetext{Equal contribution}
\urlstyle{same}
\def\thefootnote{\textdagger}\footnotetext{{Project page: \url{https://yuqunw.github.io/MonoPatchNeRF/}}}

\begin{abstract}
    The latest regularized Neural Radiance Field (NeRF) approaches produce poor geometry and view extrapolation for large scale sparse view scenes, such as ETH3D. Density-based approaches tend to be under-constrained, while surface-based approaches tend to miss details.  
    In this paper, we take a density-based approach, sampling patches instead of individual rays to better incorporate monocular depth and normal estimates and patch-based photometric consistency constraints between training views and sampled virtual views.
    Loosely constraining densities based on estimated depth aligned to sparse points further improves geometric accuracy.  While maintaining similar view synthesis quality, our approach significantly improves geometric accuracy on the ETH3D benchmark, e.g. increasing the F1@2cm score by 4x-8x compared to other regularized density-based approaches, with much lower training and inference time than other approaches.
\end{abstract}

\section{Introduction}

Modeling 3D scenes from imagery is useful for mapping, facility assessment, robotics, construction monitoring, and many other applications, which typically require both accurate geometry for measurement and realistic visualization for novel views. 

Traditional multi-view stereo~(MVS) methods predict accurate geometry given color images, but they have many limitations, such as modeling incomplete surfaces, low ability to render novel views, and reliance on complex algorithms and heuristics that inhibit further improvement. 
Neural Radiance Field~(NeRF)~\cite{mildenhall2021nerf} and 3D Gaussian Splatting~\cite{Kerbl20233DGS} provide excellent synthesis of novel views, especially when interpolating or near the training views. Recent work aims to improve the geometric accuracy in densely sampled scenes, from object-centric scale~\cite{Yariv2021VolumeRO, wang2021neus} to large building scale~\cite{yu2022monosdf, li2023neuralangelo}. However, these approaches do not perform well in large scenes with sparse views~(Fig.~\ref{fig:teaser}), which is a commonly encountered setting in many applications.  In this paper, we aim to create 3D models that provide accurate geometry and view synthesis for large scale scenes with sparse input views. 


Current approaches to improve geometric accuracy of NeRF models include guiding with monocular geometry estimates~\cite{yu2022monosdf}, applying appearance priors to virtual views~\cite{niemeyer2022regnerf}, and constraining solutions with SDF-based models~\cite{yu2022monosdf,li2023neuralangelo}. Our early experiments showed SDF-based methods have difficulty capturing details in large, complex scenes, due to sensitivity to initialization, the surface smoothness prior, and limits on volumetric resolution. 
This leads us to a density-based approach, where the challenge is regularizing an under-constrained solution space. 

The \textbf{key to our approach} is to use patch-based ray sampling to provide novel constraints that 
better incorporate monocular estimates and encourage cross-view consistency. 
Monocular depth estimators~(e.g.,~\cite{hu2024metric3dv2}) 
provide excellent estimates of local shape but may not be globally consistent. By sampling multiple rays in a local patch, we can predict depth gradients and enforce consistency up to a scale and translation.  
Also, rather than a general appearance prior~\cite{niemeyer2022regnerf}, we apply a more specific constraint of occlusion-aware patch-based photometric consistency between sampled virtual views and training views.  Density-based models tend to ``cheat'' by modeling view-dependent effects as densities near the frustrum, which we prevent with loose constraints based on monocular depth maps aligned to sparse structure-from-motion points.


Our experiments show that each of these improvements are critical to improve geometry estimates in large-scale sparse-view scenes.  Our method significantly outperforms other NeRF-based approaches on the ETH3D benchmark~\cite{schops2017multi} in geometry estimation while maintaining competitive results in novel view synthesis.  We also achieve competitive results on the Tanks and Temples benchmark~\cite{knapitsch2017tanks}, which has denser views.  Our method also has practical advantages of faster training and inference and lower memory requirements, compared to other NeRF-based approaches that aim for geometric accuracy.  Our method still falls short of the best classic MVS approaches according to point cloud metrics, but provides a good balance of geometric accuracy and novel view synthesis.

In summary, this paper offers the following main contributions:
\begin{itemize}
\item More effective use of monocular geometry estimates through patch-based ray sampling (Tab.~\ref{tab:ablation},~\ref{tab:tnt}) 
\item Effective photometric consistency constraints between training and sampled virtual views (Tab.~\ref{tab:ablation},~\ref{tab:tnt}) 
\item State-of-the-art blend of geometric accuracy and novel view synthesis for complex scenes with sparse views (Tab.~\ref{tab:eth3d}).
\end{itemize}


\section{Related Works}
\begin{figure*}[t]
\centering
\begin{tabular}{c}
        \includegraphics[width=.98\textwidth]{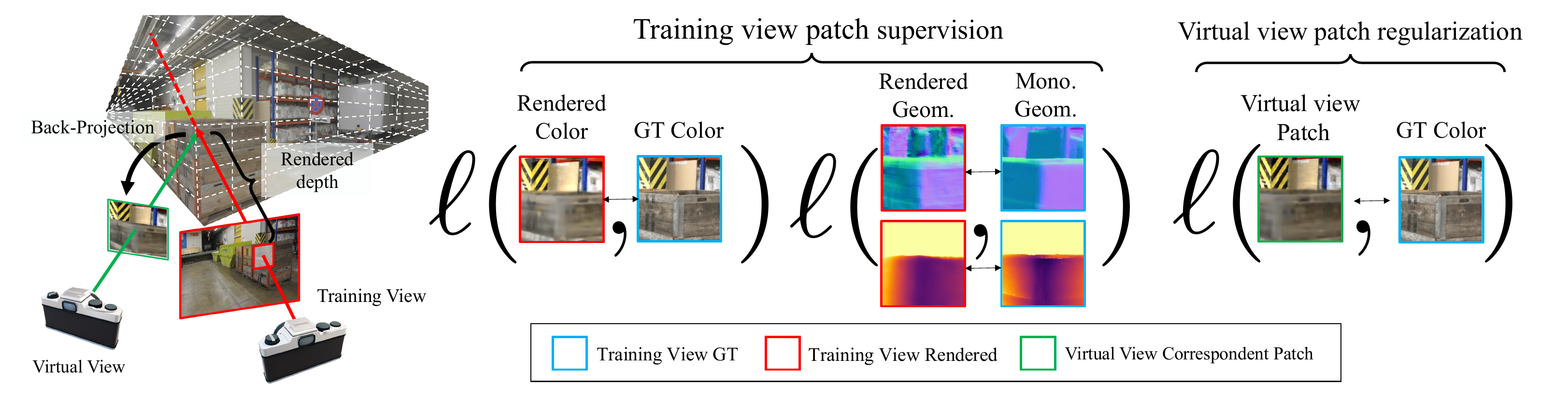}
\end{tabular}
\vspace{-0.99em}
\caption{
    \textbf{Overview of our architecture.} 
    Our MonoPatchNeRF contains three major types of losses: 1) color supervision of RGB images, 2) geometric supervision of monocular depth and normal maps, and 3) virtual view patches regularization between randomly sampled patches and corresponding ground truth RGB pixels.
    We sample the virtual view pose via random translations from the training view camera center, and obtain the virtual view corresponding patch by rendering along the back-projected ray that is unprojected with the rendered depth from the training view (Figure~\ref{fig:nvp}).
    Additionally, we limit the density search space by pruning out the regions using the monocular geometry (Figure~\ref{fig:empty_space_pruning}).
}
\label{fig:architecture}
\end{figure*}

Our work aims to create models from sparse views of complex, large-scale scenes that achieve both geometric accuracy, typically pursued by multiview stereo (MVS), and realistic novel view synthesis, as pursued by neural rendering methods.

\textbf{MVS} is a well-studied field, ranging from early works~\cite{lucas1981iterative} with pure photometric scoring to more recent approaches that incorporate learned features~\cite{yao2018mvsnet,yao2019recurrent,hanley2016aiaa,lee2021patchmatch}.
While scene representations vary, state-of-the-art methods typically predict the depth map of each image based on photometric consistency with a set of source views~\cite{schoenberger2016mvs,kuhn2020deepc,ma2021eppmvsnet,lee2021patchmatch}, and use geometric consistency across views to fuse the depths together into a single point cloud~\cite{schoenberger2016mvs,galliani2015massively}.
The fused point clouds are evaluated against the ground truth geometry for precision-recall~\cite{knapitsch2017tanks}, or accuracy-completeness~\cite{schops2017multi}, with $F_1$ score combining the harmonic mean of the two values. 
Different from MVS, our method constructs a 3D density representation and extracts depth maps using volumetric rendering of expected depth at source views, followed by a conventional depth-map fusion pipeline~\cite{galliani2015massively}.

Rather than producing precise 3D points, \textbf{neural rendering methods}, such as Neural Radiance Field~(NeRF)~\cite{mildenhall2021nerf} and 3D Gaussian Splatting (3DGS)~\cite{Kerbl20233DGS}, aim to realistically synthesize novel views by optimizing a model to render the training views. 
Many efforts boost the rendering quality~\cite{wu2022diver,barron2022mip,chen2022tensorf}, rendering efficiency~\cite{muller2022instant}, model size~\cite{lee2022qff,Rho_2023_CVPR}, and device requirements~\cite{Chen2022MobileNeRFET}. DeLiRa~\cite{Guizilini2023DeLiRaSD} incorporates multiview photometric consistency but notes that this requires overlap between input images, limiting applicability to sparsely viewed scenes. 

While NeRF and 3DGS approaches have been shown to reliably estimate geometry and appearance in dense captures~\cite{li2023neuralangelo,fan2024trim}, multiple papers~\cite{yu2021pixelnerf,jain2021putting_nerf_on_a_diet,radford_icml2021_clip,niemeyer2022regnerf,yang2023freenerf,wang2023sparsenerf} show failures in cases of outward-facing, wide-baseline, or \textbf{sparse inputs}.
PixelNeRF~\cite{yu2021pixelnerf} and DietNeRF~\cite{jain2021putting_nerf_on_a_diet} propose learning based feed-forward solutions that use prior knowledge to better handle sparse inputs.
RegNeRF~\cite{niemeyer2022regnerf} regularizes by applying appearance likelihood and geometric smoothness objectives to patches sampled from virtual views. 
SPARF~\cite{truong2023sparf} jointly optimizes the NeRF models and camera poses with extracted pixel matches on input views, and improves performance given sparse inputs.
FreeNeRF~\cite{yang2023freenerf} improves sparse-view modeling of object-scale scenes by attenuating high-frequency components of the ray positional encoding and penalizing near-field densities.
Our strategy is closest to that of RegNeRF~\cite{niemeyer2022regnerf}, as we also sample patches from virtual views to regularize appearance. We enforce a stronger constraint of photometric consistency with training views and also encourage geometric consistency with monocular estimates. 

Using image-based priors in regularization has also been shown effective in training neural SDF and NeRF-based models. 
Roessle et al.~\cite{roessle2022dense} and NerfingMVS~\cite{wei2021nerfingmvs} use a pretrained network to estimate dense depth given sparse SfM points, and supervise the NeRF model with the estimated depth.  
However, the sparse depth from SfM often contains noise that can be passed down to the dense prediction~\cite{cheng2019learning,park2020non}. 
NeuralWarp~\cite{Darmon2021ImprovingNI} proposes cross-view photometric consistency for training patches as MVS with visibility information from SfM. Our occlusion-aware photometric consistency between {\em virtual} and training views takes advantage of the novel view synthesis capability of NeRF to provide more effective constraints for sparse views.

Implicit surface models, such as \textbf{SDF}-based, model the scene in terms of distance to surfaces, rather than point densities.   MonoSDF~\cite{yu2022monosdf} guides signed distance function (SDF) models with monocular estimates of depth and normals, producing accurate surface models from videos of rooms. 
Neuralangelo~\cite{li2023neuralangelo} improves the ability of SDF-based methods to encode details such as bicycle racks and chair legs using coarse-to-fine optimization and by using numerical gradients to compute higher-order derivatives. However, our experiments show that these methods have limited effectiveness for modeling complex, large-scale scenes from sparse views.

\section{Method}

Given a collection of posed images capturing a large scale scene, our goal is to construct a 3D neural radiance field that renders high-quality images and predicts accurate and complete surface geometry. We achieve this by introducing three novel components: 1) \textbf{patch-based geometry supervision} to better leverage the local consistency of monocular cues; 2) \textbf{patch-based occlusion-aware photometric regularization} across virtual and training viewpoints, better guiding NeRF training when input views are sparse; 3) \textbf{density restriction} through monocular geometry estimates and sparse points, inhibiting NeRF from modeling view-dependent effects with densities in unlikely areas. Figure~\ref{fig:architecture} provides a summary of our approach.

\subsection{NeRF with Patch-Based Sampling}
\label{sec:nerf}

The Neural Radiance Field (NeRF)~\cite{mildenhall2020nerf} establishes a parametric representation of the scene, enabling realistic rendering of novel viewpoints from a given image collection. NeRF is defined by $\mathbf{c}, \sigma = F_\theta(\mathbf{x}, \mathbf{v})$, where $\mathbf{c}$ is output color, $\sigma$ represents opacity, $\mathbf{x}$ is the 3D point position, and $\mathbf{v}$ denotes the viewing direction. The variable $\theta$ represents learnable parameters optimized for each scene. 
The pixel color and depth can be computed through volume rendering along the corresponding ray $\mathbf{r}(t) = \mathbf{o} + t\mathbf{v}$, where $\mathbf{o}$ is the camera center and $\mathbf{v}$ is the ray direction.


NeRF is trained by minimizing the difference between the rendered RGB value $\hat{\mathbf{c}}$ and the observed value $\mathbf{c}$ of random sampled pixels across images.
Despite its exceptional performance in synthesizing nearby novel views, NeRF falls short in capturing high-quality geometry and rendering extrapolated viewpoints~(Figure~\ref{fig:eth_mesh_comp}). 
To overcome these challenges, we sample local patches instead of discrete rays, and propose our patch-based supervision and regularization.
During training, we iterate all training views, sample a batch of local $8\times8$ patches $P=\{p\}$, and apply the per-pixel Huber loss for each patch:~$L_\text{rgb} = \sum_{p \subset P} L_\text{huber}(({\mathbf{\hat{c}}_p}, \mathbf{c}_p))$.

\subsection{Distillation of Patch-based Monocular Cues}
\label{sec:monocular}

Learning-based networks for single-image normal and depth prediction can provide robust cues for the geometry, especially for areas with photometric ambiguity~(reflective and textureless surfaces). 
Taking inspiration from this, prior works~\cite{guo2022neural,yu2022monosdf} exploit monocular depth and normal supervision for neural fields, and apply a pixel-based scale and shift per batch to align monocular and predicted depth.
However, monocular methods tend to be only locally consistent.
Image-wide scale-and-shift alignment may still leave large depth errors that reduce the usefulness of monocular depth estimates for supervision.
To address this challenge, we compute the scale and shift per local patch to better leverage the capacity of monocular cues and achieve better performance~(Compare \textit{Mono.} and \textit{Mono.+Patch} in Tables~\ref{tab:ablation} and~\ref{tab:tnt}).

For each patch $P=\{p\}$, we compute the transformed monocular depth $\{ d_p^\dag \}$ using the optimal scale $s$ and shift $t$ from a least-squares criterion~\cite{manivasagam2020lidarsim} between rendered depth$\{ \hat{d}_p \}$ and monocular depth $\{ d_p \}$ on $P$.
The depth loss $L_\text{depth}$ and $L_{\nabla \text{depth}}$ are applied to penalizes the absolute and gradient discrepancy between $\{ \hat{d}_p \}$ and $\{ {d}_p^\dag \}$:

\begin{align}
L_\text{depth} &= \sum_{p \subset P} \| \hat{d}_p - {d}_p^{\dag} \| \\
L_{\nabla \text{depth}} &= \sum_{p \subset P} \| \nabla \hat{d}_p - \nabla {d}_p^{\dag} \| ,
\end{align}





Following RefNeRF~\cite{verbin2022ref}, we compute density-based normals $\mathbf{n}^\nabla_i = -\nabla \sigma_i / \|-\nabla \sigma_i\|$ with the gradient of opacity and MLP-based normals ${\mathbf{n_i}}^\theta=F_\theta(\mathbf{x_i}, \mathbf{v_i})$ with a normals rendering head. 
After volume rendering, the two normals predictions are supervised with monocular normals $\{{n}_p\}$ using angular and $L_1$ loss:

\begin{equation}
\begin{split}
L_\text{normal} &= \sum_{p \subset P} (1 - \cos({\mathbf{n}}_p, \mathbf{n}^\nabla_p) + |{\mathbf{n}}_p - \mathbf{n}^\nabla_p|) \\
&+ \sum_{p \subset P} (1 - \cos({\mathbf{n}}_p, \mathbf{n}^\theta_p) + |{\mathbf{n}}_p - \mathbf{n}^\theta_p|)
\end{split}
\end{equation}

We also apply the gradient loss $L_{\nabla \text{normal}}$ over $\mathbf{n}^\nabla$: 
\begin{equation}
\begin{split}
L_{\nabla \text{normal}} &= \sum_{p \subset P} (|\nabla {\mathbf{n}}_p - \nabla \mathbf{n}^\nabla_p|)
\end{split}
\end{equation}

Finally, our patch-based monocular loss is defined as: 
\begin{equation}
    L_\text{mono} = L_\text{depth} + L_{\nabla \text{depth}} + L_\text{normal} + L_{\nabla \text{normal}}.
\end{equation}

\subsection{Patch-based Photometric Consistency over Virtual Views}
\label{sec:regularization}


To better train NeRF models given sparse input views, we enforce an occlusion-aware photometric consistency between training patches and randomly sampled virtual patches.
Though designed for sparse-view input, we find that the photometric regularization works for general setups, significantly boosting the performance for both sparse-view~(Table~\ref{tab:ablation}) and dense-view scenes~(Table~\ref{tab:tnt}).




\begin{figure}
\centering
\begin{tabular}{l}
    \includegraphics[trim=1.5cm 0cm 0cm 0cm, clip, width=0.49\textwidth]{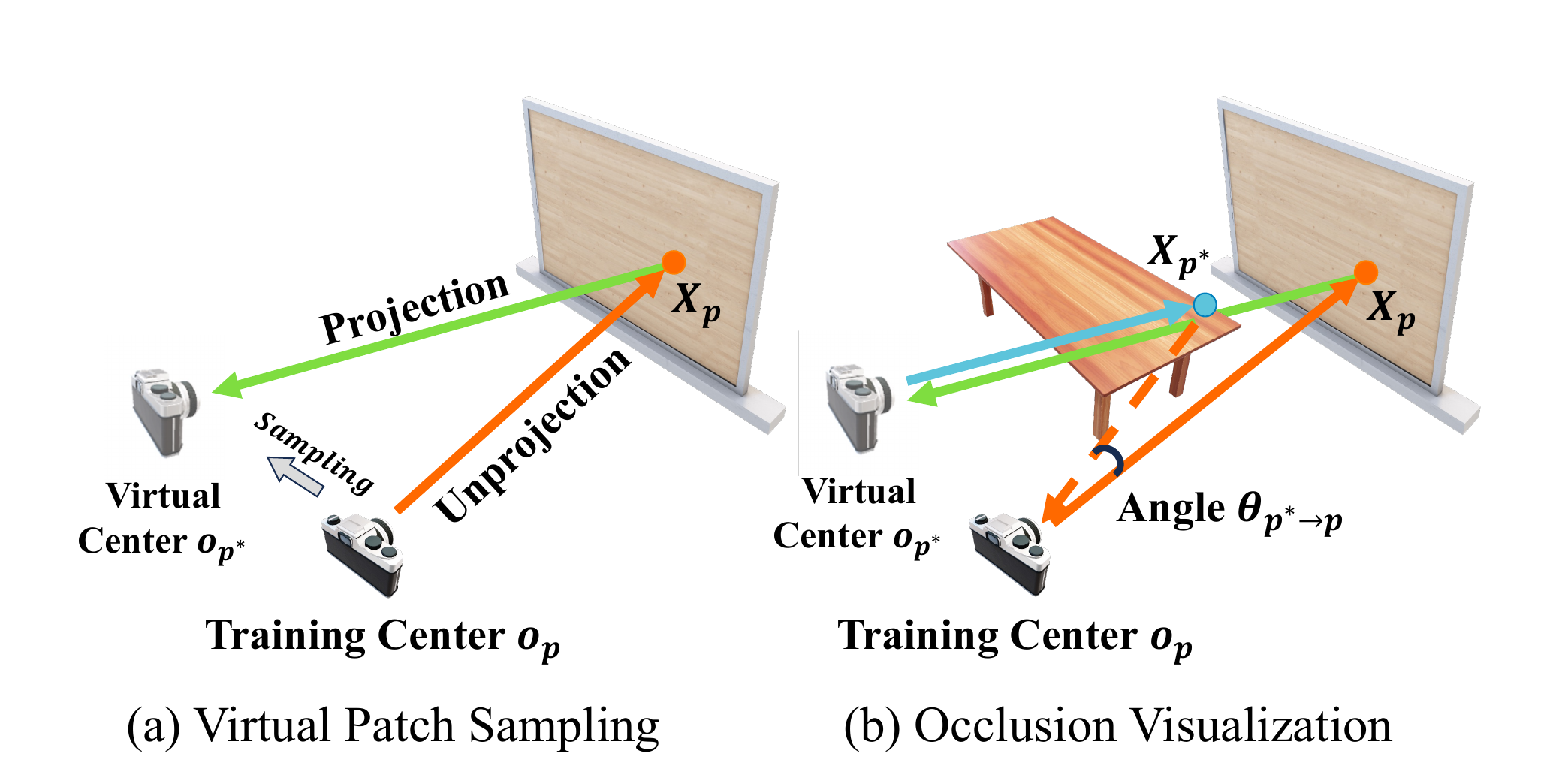} \\
     \vspace{-8mm}

\end{tabular}
\caption{
    \textbf{Virtual view patch sampling and occlusion visualization.} 
    (a) We first sample a virtual center $\textbf{o}_{p^\ast}$ near the training center $\textbf{o}_{p}$. We then unproject the training patch to $\{\textbf{X}_p\}$ with rendered depth, and project $\{\textbf{X}_p\}$ to $\textbf{o}_{p^\ast}$ for the virtual patch viewing directions. Color of the virtual patch are rendered and compared to the ground truth RGB in the training patch. 
    (b) We unproject the virtual patch to $\{\textbf{X}_{p^\ast}\}$ with virtual rendered depth, and mask pixels based on the angle $\{\theta_{p^\ast \rightarrow p}\}$ between $\{\textbf{X}_p\}$ to $\textbf{o}_p$ and $\{\textbf{X}_{p^\ast}\}$ to $\textbf{o}_p$.
    For simplicity, the visualization only contains a single pixel $p$. 
}

\vspace{-1em}
\label{fig:nvp}
\end{figure}

As visualized in Fig.~\ref{fig:nvp}, for each training patch $P$, we randomly sample a corresponding virtual camera center $o_{p^\ast}$ within a fixed distance to $o_p$~($0.05 \times$ the scene width). 
The correspondence is determined by unprojecting the pixel from the training view into 3D points, denoted as $\mathbf{X}_p = \mathbf{o}_p + \hat{d}_p\mathbf{v}_p$, and then back-projecting $\mathbf{X}_p$ into the $\mathbf{o}_{p^\ast}$ for the corresponding direction $\mathbf{v}_{p^\ast} = (\mathbf{X}_p - \mathbf{o}_{p^\ast}) / \|(\mathbf{X}_p - \mathbf{o}_{p^\ast})\|$. 
Given $\mathbf{o}_{p^\ast}$ and $\mathbf{v}_{p^\ast}$, we render the color $\mathbf{\hat{c}}_{p^\ast}$ and depth ${\hat{d}}_{p^\ast}$.
Since viewing the same point from different perspectives can result in varying occlusion patterns, we additionally apply mask $M_{p\rightarrow p^\ast}$ to remove any clearly occluded pixels to eliminate the occlusion impact.  
We compute $M_{p\rightarrow p^\ast}$ by comparing an angle threshold $\theta_{\text{thresh}}$ and the angle $\theta_{p^\ast \rightarrow p}$ between rays from $\mathbf{o}_p$ to $\mathbf{X}_p$ and $\mathbf{o}_p$ to $\mathbf{X}_{p^\ast}$:
\begin{equation}
\begin{split}
\theta_{p^\ast \rightarrow p} &= \arccos(\mathbf{v}_{p^\ast \rightarrow p} \cdot \mathbf{v}_p), \\
M_{p\rightarrow p^\ast} &= 
  \begin{cases} 
   1 & \text{if } \theta_{p^\ast \rightarrow p} \leq \theta_{\text{thresh}} \\
   0 & \text{otherwise}
  \end{cases} \\
\end{split}
\end{equation}
where $\mathbf{v}_{p^\ast \rightarrow p} = \frac{(\mathbf{X}_{p^\ast}  - \mathbf{o}_{p})}{\|(\mathbf{X}_{p^\ast}  - \mathbf{o}_{p})\|}$ and $
\mathbf{X}_{p^\ast} = \mathbf{o}_{p^\ast} + \hat{d}_{p^\ast}\mathbf{v}_{p^\ast}$.

The occlusion-aware patch-based consistency loss is defined as:
\begin{equation}
\begin{split}
L_\text{virtual} &= L_\text{SSIM}(\{\mathbf{\hat{c}}_{p^\ast}\}, \{{\mathbf{c}}_p\}, \{M_{p\rightarrow p^\ast}\}) \\
&+ L_\text{NCC}(\{\mathbf{\hat{c}}_{p^\ast}\}, \{{\mathbf{c}}_p\}, \{M_{p\rightarrow p^\ast}\})
\end{split}
\end{equation}

where $L_\text{SSIM}$ and $L_\text{NCC}$ measure the structural similarity and normalized cross-correlation between two masked patches respectively.
Both losses are robust to view-dependent effects like illumination change.
Though we rely on rendered depth for the loss, the occlusion mask will inhibit the loss from incorrectly guiding the optimization in the beginning.

\subsection{Density Restriction by Empty Space Pruning with Sparse SfM Geometry}
\label{sec:pruning}
\begin{figure}
\centering
\includegraphics[width=0.47\textwidth]{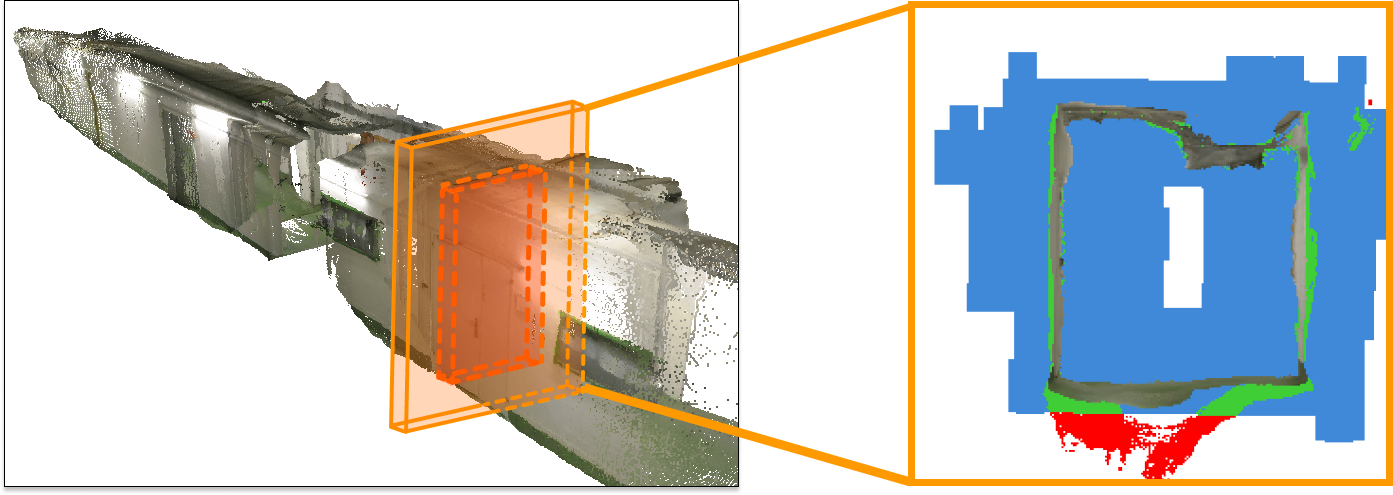}
\captionof{figure}{
    \textbf{Visualization of density restrictions.} On the left, we present the point cloud reconstruction of our model trained with density restrictions. On the right, a vertical slice of the reconstructed scene is shown, both with and without density restrictions. The original scene points and color points (green and red) represent our reconstructed point cloud with and without density restrictions, respectively. The blue area denotes density-restricted voxels. With density restrictions, the ground is accurately reconstructed as a plane, whereas without density restrictions, the ground sinks down.
}
\vspace{-1em}
\label{fig:empty_space_pruning}
\end{figure}

One significant challenge in NeRF is the occurrence of floaters and background collapse~\cite{barron2022mip}.
The challenge arises because NeRF fails to predict correct geometry for surfaces with low texture and view-dependent effects, or tends to overfit in near-camera regions that are unseen from other views during training.  
We address this problem by limiting the domain of density distributions using monocular geometric prior and sparse multi-view prior~(Figure~\ref{fig:empty_space_pruning}).

Monocular depth provides useful relative distance information, and SfM points provide metric depth that can be utilized for aligning the monocular depth with 3D space. 
For each view, we use RANSAC to solve a scale and shift for monocular depth with the projected sparse points to minimize the influence of noise in the points. 
We then constrain the grid density distribution to a specified interval around the estimated depth along each ray. 
This hard restriction effectively prunes out empty space, thereby eliminating the floaters and improving the overall geometry estimation.

\subsection{Training}
\label{sec:training}
We start by estimating monocular geometric cues with images using the pretrained Omnidata~\cite{eftekhar_iccv2021_omnidata} model. 
With the sparse points from SfM, we use RANSAC to find the optimal shift and scale for the monocular depth for density restriction. 
To initialize the density restriction, we voxelize the space, project the center of each voxel to all training views, label voxels with centers lying within 20\% of any monocular depth map, and exclude sampling outside of labeled voxels.
During training, we sample $128$ patches per iteration, sample one virtual patch for each training patch, and evaluate the loss terms for all patches.
Angle threshold $\delta_{\text{thresh}}$ is set as 10 degrees when estimating occlusion masks.
We use NerfAcc~\cite{li2022nerfacc} with modified QFF~\cite{lee2022qff} as our base model for faster training and inference without loss of accuracy, and train with the unified loss $L = \sum \lambda_{:} L_{:}$, with $\lambda_{\text{rgb}}=1.0$, $\lambda_{\text{depth}}=0.05$, $\lambda_{\nabla \text{depth}}=0.025$, $\lambda_{\text{normal}}=1\times10^{-3}$, $\lambda_{\nabla \text{normal}}=5\times10^{-4}$, $\lambda_{\text{SSIM}}=1\times10^{-4}$, $\lambda_{\text{NCC}}=1\times10^{-4}$. 
Please see our supplementary material for more details on parameters and model architecture.

\input{figures/eth3d}
\section{Experiments}

Our experiments investigate: (1) the geometric accuracy and rendering quality of existing NeRF methods and our method on the challenging MVS benchmarks, as measured by point cloud metrics and novel view synthesis; (2) how each of our contributions and system components affect performance.

\noindent\textbf{Datasets}: 
We experiment with ETH3D~\cite{schops2017multi} and Tanks and Temples (TnT)~\cite{knapitsch2017tanks} because they are among the most challenging benchmarks for MVS.  We use the training scenes from the ETH3D High-Resolution dataset~(ETH3D) ~\cite{schops2017multi}, consisting of 7  large-scale indoor scenes and 6 outdoor scenes. These scenes are sparsely captured, with  only 35 images on average, which is especially challenging for NeRF. In addition, we experiment with TnT large-scale indoor scenes~(\textit{Church}, \textit{Meetingroom}) and outdoor scenes~(\textit{Barn}, \textit{Courthouse}), as well as the advanced testing scenes used in the~\cite{yu2022monosdf} to validate our method on densely captured scenes.

\noindent\textbf{Evaluation Protocols}:
We evaluate the methods on novel view synthesis and geometric inference. 
For novel view synthesis, we train the model with 90\% of the images and treat the remaining 10\% images as test views for evaluation. We report Peak Signal-to-Noise Ratio~(PSNR), Structural Similarity Index Measure~(SSIM), and Learned Perceptual Image Patch Similarity~(LPIPS)~\cite{zhang2018perceptual} as evaluation metrics. 
For geometric inference, we train on all images, and evaluate using the provided 3D geometry evaluation pipeline~\cite{schops2017multi, knapitsch2017tanks} on point clouds. 
To generate point clouds, for density-based methods, we render the expected depth map of each training view, and fuse the depth maps with the scheme proposed by Galliani et al~\cite{galliani2015massively}; for SDF-based methods, we render the mesh with SDF values using marching cube~\cite{lorensen1987marching} and sample points from the rendered mesh. For TnT dataset, we only report the geometry inference as all views are densely captured. Since NeRF papers rarely report F1-scores, we train and evaluate scene models for some of the leading methods in density-based and SDF-based NeRF, using author-provided code, advice, and reported numbers where possible.

\begin{figure}[t]
\centering
\begin{tabular}{l}
        {\includegraphics[trim=0.84cm 0cm 0cm 0cm, clip, width=0.463\textwidth]{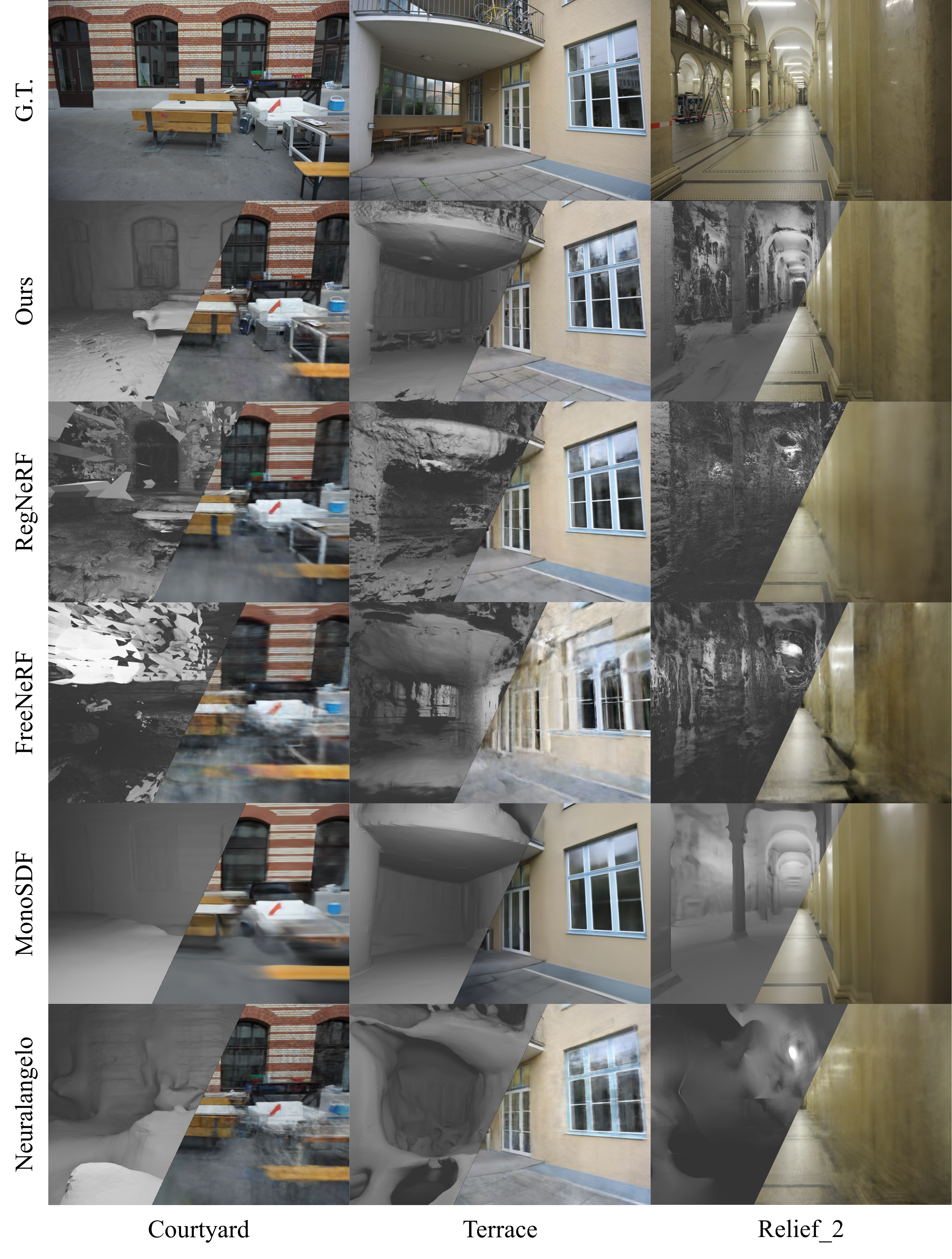}}
\end{tabular}
\vspace{-1.5em}
\caption{\textbf{Qualitative comparison of novel view images and meshes.} We provide test view rendered images and meshes on the ETH3D dataset~\cite{schops2017multi}. The mesh of Ours, RegNeRF~\cite{niemeyer2022regnerf} and FreeNeRF~\cite{yang2023freenerf} are generated via TSDF fusion given predicted RGBD sequence. Best viewed when zoomed in.}
\vspace{-1.5em}
\label{fig:eth_mesh_comp}
\end{figure}


\begin{table*}[h]
\caption{
    \textbf{Quantitative evaluation on ETH3D~\cite{schops2017multi}}. 
    We report baselines, our results with and without MVS depth based guidance, and reference MVS results on ETH3D~\cite{schops2017multi}. 
    We denote \texttt{NVS} as the model's ability to perform novel-view synthesis, and the indoor and outdoor scenes in the ETH3D dataset as \texttt{In}, \texttt{Out}.
    The top rows show the baselines and our methods without using additional multi-view supervision, 
    and the bottom rows show the reference MVS results and our method supervised with ACMMP~\cite{Xu2022Multi} depth.
    We use author provided codes to evaluate the baselines, and ETH3D webpage provided results for MVS.
    We mark the top methods in blue and green. ($\fboxsep=0pt\fbox{\color{ours_text_blue}\rule{2.5mm}{2.5mm}}$ best,
    $\fboxsep=0pt\fbox{\color{ours_green}\rule{2.5mm}{2.5mm}}$ second best)}
\vspace{-1.0em}
\centering
\resizebox{1.0\textwidth}{!}{
\begin{tabular}{l|cccc|c@{\hspace{2mm}}c@{\hspace{2mm}}c@{\hspace{2mm}}c@{\hspace{2mm}}c@{\hspace{2mm}}c}
\toprule
\multirow{2}{*}{Method} &\multirow{2}{*}{NVS} & \multirow{2}{*}{PSNR $\uparrow$} & \multirow{2}{*}{SSIM $\uparrow$} & \multirow{2}{*}{LPIPS$\downarrow$} & Prec.$_{2cm}$ $\uparrow$ & Recall$_{2cm}$ $\uparrow$ & F-score$_{2cm}$ $\uparrow$ & F-score$_{5cm}$ $\uparrow$ & Time~(hrs/secs) $\downarrow$ & \multirow{2}{*}{Step} \\
&&&&& \small{Mean / In/ Out} &  \small{Mean / In / Out} & \small{Mean / In / Out} & \small{Mean / In / Out} & \small{Training / Inference}\\
\midrule
RegNeRF~\cite{niemeyer2022regnerf}          & \cmark & \fc{20.90} & \scolor{0.707} & 0.439 &  7.3	/ 11.1 / 2.8 & 6.0 / 9.5	/ 1.9 & 6.4 / 10.0 / 2.2  & 15.5 / 22.4 / 7.4  & 14.4 / 36.4 & 200000 \\
FreeNeRF~\cite{yang2023freenerf}            & \cmark & 17.24 & 0.590 & 0.581 & 7.5 / 10.4 / 4.1 & 2.6 / 3.8 / 1.3 & 3.3 / 4.7 / 1.7 & 8.5 / 10.8 / 5.7 & \scolor{7.0 / 30.4}  & 200000 \\
MonoSDF~\cite{yu2022monosdf}               & \cmark & 18.85 & 0.679 & 0.498 & \scolor{25.2} / \scolor{26.2} / \scolor{24.0} & \scolor{19.3} / \scolor{28.5} / \scolor{8.5} & \scolor{20.1} / \scolor{26.9} / \scolor{12.1} & \scolor{41.1} / \scolor{45.2} / \scolor{36.4} & 20.9 / 136.0 & 200000 \\
Neuralangelo~\cite{li2023neuralangelo}      & \cmark & 19.53 & 0.696 & \scolor{0.414} & 3.3 / 3.4 / 3.2 & 2.1 / 3.4 / 0.6 & 2.3 / 3.4 / 1.0 & 7.2 / 8.0 / 6.2 & 19.9 / 61.9 & 200000 \\
Ours                                      & \cmark & \scolor{20.12} & \fc{0.720} & \fc{0.379} &  \fc{36.2} / \fc{45.6} / \fc{25.2} & \fc{{24}.4} / \fc{29.8} / \fc{{18}.2} & \fc{28.8} / \fc{35.6} / \fc{20.9} & \fc{46.9} / \fc{52.9} / \fc{40.0} & \fc{2.2 / 4.6} & 50000 \\
\midrule
Ours (MVS-Depth)            & \cmark & {20.48} & {0.742} & {0.341}  & {70.2} / 71.6 / 68.4 & {53.6} / 58.5 / 47.9 & {60.4} / 64.0 / 56.3 & {80.7} / 81.7 / 75.3 & 2.7 / 4.6 & 50000 \\
\midrule
Gipuma~\cite{galliani2015massively}         & \xmark & - & - & -     & 86.5 / 89.3 / 83.2           & 24.9 / 24.6 / 25.3 & 36.4 / 35.8 / 37.1 & 49.2 / 47.1 / 51.7 & - / - & -\\
COLMAP~\cite{schoenberger2016mvs}           & \xmark & - & - & -     & \fc{91.9} / {95.0} / 88.2   & 55.1 / 52.9 / 57.7 & 67.7 / 66.8 / 68.7 & 80.5 / 78.5 / 82.9 & - / - & - \\
ACMMP~\cite{Xu2022Multi}                    & \xmark & - & - & -     & 90.6 / 92.4 / {88.6}       & \fc{77.6} / \fc{79.6} / \fc{75.3} & \fc{83.4} / \fc{85.3} / \fc{81.3} & \fc{92.0} / \fc{92.2} / \fc{91.9} & - / - & -\\
\bottomrule
\end{tabular}
}
\label{tab:eth3d}
\end{table*}

\subsection{Main Results}

\begin{figure}[h]
\centering
\begin{tabular}{l}
        {\includegraphics[trim=0.7cm 0cm 0cm 0cm, clip, width=0.47\textwidth]{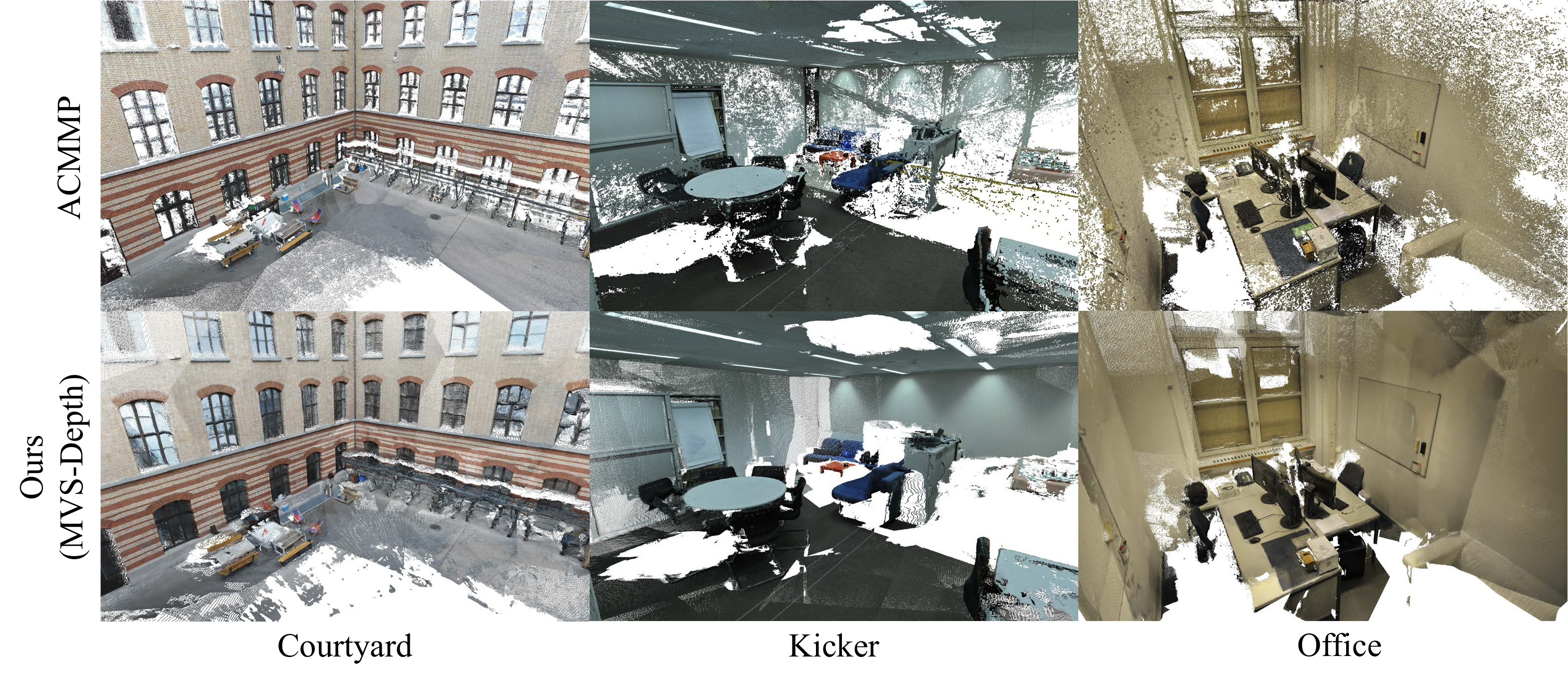}}    
\end{tabular}
\vspace{-1.5em}
\caption{\textbf{Point cloud visualization.} We visualize the point clouds of ACMMP~\cite{Xu2022Multi} and our method with MVS depth supervision on ETH3D~\cite{schops2017multi}. Our method 
is able to complete textureless and reflective surfaces. }
\label{fig:pc_ours_mvs}
\vspace{-0.5em}
\end{figure}

Table~\ref{tab:eth3d} compares our approach to density-based and SDF-based NeRF approaches and to MVS for both novel view synthesis and point cloud accuracy metrics. Fig.~\ref{fig:eth3d_qualitative} compares RGB, depth, and normal renderings of the NeRF-based approaches, and Fig.~\ref{fig:eth_mesh_comp} shows more examples of meshes and novel view synthesis. Fig.~\ref{fig:pc_ours_mvs} compares generated point clouds to MVS. Table~\ref{tab:tnt_comp} compares geometric accuracy to SDF-based methods on the TnT dataset. More results like novel view video comparisons and sparse-view TnT comparisons are in the supplemental.



\noindent \textbf{Comparison to regularized density-based NeRF}: Our approach provides more accurate and detailed novel view synthesis, depth, normals, and meshes than RegNeRF~\cite{niemeyer2022regnerf} and FreeNeRF~\cite{yang2023freenerf} in the ETH3D experiments (Tab.~\ref{tab:eth3d}, Fig.~\ref{fig:eth3d_qualitative}, Fig.~\ref{fig:eth_mesh_comp}).  Our approach particularly outperforms on texture-less, semi-transparent, or reflective surfaces (e.g., tables, glass doors, and windows). RegNeRF is second best and is closest to our approach. While RegNeRF samples patches in virtual views to regularize based on their color and geometry likelihood, our key differentiation is encouraging patch-based geometric consistency with monocular cues and patch-based photometric consistency with virtual views. Our ablation (Table~\ref{tab:ablation}) confirms that without using patches or without incorporating both of these types of consistencies, our method achieves F-scores more similar to these others.

\begin{table}[b]
\vspace{-1em}
\caption{
    \textbf{Comparisons on TnT~\cite{knapitsch2017tanks}}. 
    We report the $F$-score of each method on Tanks and Temples Dataset. 
    $^\ast$ indicates results from Neuralangelo~\cite{li2023neuralangelo}, acquired with scale initialization from ground truth points. $^{\text{\textdagger}}$ indicates results from MonoSDF~\cite{yu2022monosdf}. For ease of comparison against Neuralangelo, we exclude scene \textit{Church} which is not provided by the authors for comparison.
    \textit{NeuralA.} refers to \textit{Neuralangelo}. 
    Advanced scene results are from the official online evaluation site.
    We mark the best scoring methods with \textbf{bold}. 
}
\vspace{-2.em}
\begin{center}
\resizebox{0.48\textwidth}{!}{
\begin{tabular}{c|c|cccc|c|c|cc}
\toprule
                          &\sm{Scene}       & \sm{Ours}         & \sm{MonoSDF}  & \sm{NeuralA.$^\ast$} & \sm{NeuralWarp$^\ast$} & & \sm{Scene} & \sm{Ours} & \sm{MonoSDF$^{\text{\textdagger}}$}  \\
\midrule
\multirow{5}{*}{\rotatebox[origin=c]{90}{\parbox[c][4mm][c]{13mm}{Training}}} & \sm{Meetingroom} & 22.0         & 27.2  & \textbf{32.0} & 8.0  & 
\multirow{5}{*}{\rotatebox[origin=c]{90}{\parbox[c][4mm][c]{15mm}{Advanced}}} & \sm{Auditorium} & \textbf{8.0} & 3.2  \\
                          & \sm{Barn}        & 49.4         & 6.0  & \textbf{70.0} & 22.0  &    
                          & \sm{Ballroom} & \textbf{26.6} & 3.7  \\
                          & \sm{Courthouse}  & \textbf{38.3}& 6.1  & 28.0 & 8.0  &             
                          & \sm{Courtroom} & \textbf{17.2} & 13.8 \\ 
                          & \sm{Church}    & 20.3         & \textbf{21.8}  & -  & -  &               
                          & \sm{Museum} & \textbf{21.4} & 5.7  \\ 
\cmidrule{2-6}
\cmidrule{8-10}
                          & \sm{Mean}        & 36.6    & 13.1    & \textbf{43.3} & 12.7 & 
                          & \sm{Mean} & \textbf{18.3} & 6.5 \\
\bottomrule
\end{tabular}
}
\end{center}
\label{tab:tnt_comp}
\vspace{-1.5em}
\end{table}

\noindent \textbf{Comparison to SDF-based approaches}: SDF-based approaches like MonoSDF~\cite{yu2022monosdf}, Neuralangelo~\cite{li2023neuralangelo}, and  NeuralWarp~\cite{Darmon2021ImprovingNI} are well-regularized by solving directly for an implicit surface, and more easily incorporate monocular geometry estimates. SDF-based approaches work very well in some scenes, but have drawbacks of higher dependence on initialization, compute-heavy optimization, and overly smooth surfaces in the results. Qualitative results in the ETH3D scenes (Fig.~\ref{fig:eth3d_qualitative}) show that our method better captures details, such as the stairs, lamppost, and tables. In Tab.~\ref{tab:eth3d}, our method outperforms MonoSDF and Neuralangelo in all novel view synthesis and geometry metrics.  Interestingly, Neuralangelo outperforms MonoSDF in NVS but greatly underperforms in geometric accuracy because the sparse views do not provide sufficient regularization to constrain geometry. 

The denser views and often simpler scenes make TnT scenes more amenable to SDF-based methods (Tab.~\ref{tab:tnt_comp}). In the simpler Training scenes, our method performs best for ``Courthouse'', while Neuralangelo performs best for ``Meetingroom'' and ``Barn'' and MonoSDF slightly outperforms ours in ``Church''. For Advanced TnT scenes, our method consistently outperforms MonoSDF.

\begin{table}[t]
\caption{
    \textbf{Ablations on ETH3D~\cite{schops2017multi}}. 
    We report evaluations and training time~(seconds per 1000 steps) of different combinations of our components on the ETH3D dataset~\cite{schops2017multi}. 
    \texttt{Patch} denotes using patch-based training, \texttt{Mono.} denotes using monocular geometric cues, \texttt{Virtual} denotes using virtual view-based regularization, and \texttt{Restr.} denotes density restriction.
    We mark the best-performing combinations for each criterion in \textbf{bold}.
}
\centering
\resizebox{0.49\textwidth}{!}{
\begin{tabular}{cccc|ccc|c@{\hspace{3mm}}c@{\hspace{3mm}}|c}
\toprule
{Patch} & {Mono.} & {Virtual} &{Restr.} & {PSNR $\uparrow$} & {SSIM $\uparrow$} & {LPIPS$\downarrow$} & F-1$_{2cm}$ $\uparrow$ & F-1$_{5cm}$ $\uparrow$ & {Time $\downarrow$} \\
\midrule
 &  &  &  & 18.8 & 0.695 & 0.397 &
6.2 &
14.2 &
129 \\
 & \cmark &  &  & 19.6 & 0.695 & 0.393 & 
6.7 &
15.1 & 
150 \\
 &  &  & \cmark & 18.2 & 0.618 & 0.484 & 
10.3 &
23.1 & 
97 \\
\cmark & \cmark &  &  &20.0 & 0.723 & 0.388 & 
11.7 & 
24.0 & 
130 \\
\cmark &  & \cmark &  & \f{21.4} & \f{0.745} & 0.382 &
18.3 &
33.7 & 
165 \\
\cmark & \cmark & \cmark &  & 20.9 & 0.742 & \f{0.372} & 
23.1 &
38.7 & 
233 \\
\cmark & \cmark & \cmark & \cmark & 20.1 & 0.720 & 0.379 & 
\f{28.8} & 
\f{46.9} & 
153 \\
\bottomrule
\end{tabular}
}
\label{tab:ablation}
\end{table}

Robustness and compute can also be a concern for SDF-based methods. For example, Neuralangelo requires scale initialization from ground truth point clouds. For MonoSDF, following author advice, we find that reducing the bias parameter (initialized sphere radius) from defaults gives much better results. With this setting, MonoSDF captures the columns in \textit{Relief\_2} (Fig.~\ref{fig:eth_mesh_comp}) but still does not extend to the full depth of the gallery. Neuralangelo is memory-intensive; we had to reduce the batch size from default to train using one A40 GPU (48GB). By contrast, our method uses the same parameters for all experiments, and scale is set automatically based on SfM sparse points, which are typically available for posed images. Our memory and compute requirements are lower. 
See Tab.~\ref{tab:eth3d} and supplemental for details.

\noindent \textbf{Comparison to MVS}: MVS methods, such as ACMMP~\cite{Xu2022Multi}, produce very accurate geometry in some portions of scenes but also tend to produce noisy points and incomplete surfaces, and they cannot synthesize realistic novel views.  Quantitatively, our method underperforms MVS according to point cloud metrics (Table~\ref{tab:eth3d}), but qualitatively (Fig.~\ref{fig:pc_ours_mvs}) our method's point cloud is more complete and similarly accurate, especially when using guidance from MVS depth maps (described in Sec.~\ref{sec:exp_ablation}).  Much of the gap is likely due to limitations in extracting the 3D point clouds from the NeRF model. We use a simple approach of rendering expected depth in each view and applying standard fusion techniques as a post-process, while MVS methods often include optimization steps to improve geometric consistency across views. Another possible cause is that multiscale MVS methods better exploit high resolution images.

\begin{table}[t]
\caption{
    \textbf{Ablations on TnT~\cite{knapitsch2017tanks} dataset Training scenes}. 
    We show evaluations of different configurations on the selected TnT scenes. We report the $F$-score for each scene and mark the best performing configuration in \textbf{bold}.
}
\centering
\vspace{-0.25em}
\resizebox{0.49\textwidth}{!}{

\begin{tabular}{cccc|cccc|c}
\toprule
{{Patch}} & {{Mono.}} & {{Virtual}} & {{Restr.}} &
{Church} & {Meetingroom} & {Barn} & {Courthouse}&  {{Mean} }\\
 
\midrule
 &  &  &  &  1.6 &  4.7 & 16.1 &  4.5 &  6.7  \\
 & \cmark &  &  & 1.7 & 7.7 & 22.9 & 6.2 & 9.6  \\
 &  &  & \cmark & 10.2 & 10.6 & 24.7 & 16.4 & 15.5  \\
\cmark &  & \cmark &  &  7.8 & 13.0 & 35.7 & 11.9 & 17.1  \\
\cmark & \cmark &  &  &  2.3 & 13.7 & 38.0 & 22.4 & 19.1  \\
\cmark & \cmark & \cmark & \cmark & \textbf{20.3} & \textbf{22.0} & \textbf{49.4} & \textbf{38.3} & \textbf{32.5}  \\
\bottomrule
\end{tabular}
}
\label{tab:tnt}
\vspace{-0.5em}
\end{table}

\subsection{Ablation Study}
\label{sec:exp_ablation}

\noindent \textbf{Key Contributions}:
We ablate each key contribution of our approach in Tables~\ref{tab:ablation} and \ref{tab:tnt}.
Patch-based objectives, monocular cues, virtual view-based regularization, and density restriction are all important for geometry estimation.  Table~\ref{tab:ablation} indicates that monocular cues and the density restriction very slightly decrease the novel view synthesis quality, likely because they prevent the model from using erroneous geometry to create some view-dependent effects that it otherwise has trouble modeling. When using monocular supervision without patch-based training, loss for gradient of depth $L_{\nabla \text{depth}}$ and normals $L_{\nabla \text{normal}}$ are not applied as gradients are less accurate among randomly sampled pixels.

\noindent \textbf{Incorporating MVS Depth}:
We investigate how improved depth maps can affect results of our method. We use ACMMP~\cite{Xu2022Multi} inferred depth (denoted as $\hat{d}_p^{mvs}$ for simplicity) to supervise our rendered depth with an additional L1 Loss $L_\text{mvs} = |\hat{d}_p^{mvs} - {d}_p|$ and weight $\lambda_\text{mvs} = 0.1$. We do not apply scale and shift for $\hat{d}_p^{mvs}$ because MVS depth is metric depth.
In Table~\ref{tab:eth3d}, ``Ours (MVS-Depth)'' shows that these losses based on MVS depth significantly boost the geometry and slightly improve the rendering. 


\section{Conclusion}
We propose MonoPatchNeRF, a patch-based regularized NeRF model that aims to produce geometrically accurate models.
We demonstrate the effective use of monocular geometry estimates with patch-based ray sampling optimization and density constraints, as well as the effectiveness of NCC and SSIM photometric consistency losses between patches from virtual and training views. 
Our method significantly improves geometric accuracy, ranking top in terms of $F_1$, SSIM, and LPIPS compared to state-of-the-art regularized NeRF methods on the challenging ETH3D MVS benchmark.  Still, there are many potential directions for improvement, including: guided sampling of virtual view patches; joint inference of geometry with single-view predictions; including per-image terms to better handle lighting effects; expanding material models, e.g. with both diffuse and specular terms per point; incorporating semantic segmentation; and reducing memory and computation requirements.


\noindent \textbf{Acknowledgement}
This work is supported in part by NSF IIS grants 2312102 and 2020227. S.W. is supported by NSF 2331878 and 2340254, and research grants from Intel, Amazon, and IBM. Thanks to Zhi-Hao Lin for sharing the code for pose interpolation and video generation, and Bowei Chen and Liwen Wu for proofreading the paper.
{
    \small
    \bibliographystyle{splncs04}
    \bibliography{bib_merged}

\begin{thebibliography}{10}
\providecommand{\url}[1]{\texttt{#1}}
\providecommand{\urlprefix}{URL }
\providecommand{\doi}[1]{https://doi.org/#1}

\bibitem{barron2022mip}
Barron, J.T., Mildenhall, B., Verbin, D., Srinivasan, P.P., Hedman, P.: Mip-nerf 360: Unbounded anti-aliased neural radiance fields. In: CVPR (2022)

\bibitem{chen2022tensorf}
Chen, A., Xu, Z., Geiger, A., Yu, J., Su, H.: Tensorf: Tensorial radiance fields. In: European Conference on Computer Vision (ECCV) (2022)

\bibitem{Chen2022MobileNeRFET}
Chen, Z., Funkhouser, T.A., Hedman, P., Tagliasacchi, A.: Mobilenerf: Exploiting the polygon rasterization pipeline for efficient neural field rendering on mobile architectures. 2023 IEEE/CVF Conference on Computer Vision and Pattern Recognition (CVPR) pp. 16569--16578 (2022)

\bibitem{cheng2019learning}
Cheng, X., Wang, P., Yang, R.: Learning depth with convolutional spatial propagation network. IEEE transactions on pattern analysis and machine intelligence  \textbf{42}(10),  2361--2379 (2019)

\bibitem{Darmon2021ImprovingNI}
Darmon, F., Bascle, B., Devaux, J.C., Monasse, P., Aubry, M.: Improving neural implicit surfaces geometry with patch warping. In: CVPR. pp. 6250--6259 (2022)

\bibitem{eftekhar_iccv2021_omnidata}
Eftekhar, A., Sax, A., Bachmann, R., Malik, J., Zamir, A.: Omnidata: A scalable pipeline for making multi-task mid-level vision datasets from 3d scans. In: ICCV (2021)

\bibitem{fan2024trim}
Fan, L., Yang, Y., Li, M., Li, H., Zhang, Z.: Trim 3d gaussian splatting for accurate geometry representation. arXiv preprint arXiv:2406.07499  (2024)

\bibitem{galliani2015massively}
Galliani, S., Lasinger, K., Schindler, K.: Massively parallel multiview stereopsis by surface normal diffusion. In: Proceedings of the IEEE International Conference on Computer Vision. pp. 873--881 (2015)

\bibitem{Guizilini2023DeLiRaSD}
Guizilini, V.C., Vasiljevic, I., Fang, J., Ambrus, R., Zakharov, S., Sitzmann, V., Gaidon, A.: Delira: Self-supervised depth, light, and radiance fields. ICCV pp. 17889--17899 (2023)

\bibitem{guo2022neural}
Guo, H., Peng, S., Lin, H., Wang, Q., Zhang, G., Bao, H., Zhou, X.: Neural 3d scene reconstruction with the manhattan-world assumption. In: IEEE Conference on Computer Vision and Pattern Recognition (CVPR) (2022)

\bibitem{hanley2016aiaa}
Hanley, D., Bretl, T.: An improved model-based observer for inertial navigation for quadrotors with low cost imus. In: to appear in AIAA Guidance, Navigation, and Control Conference (AIAA-GNC) (2016)

\bibitem{hu2024metric3dv2}
Hu, M., Yin, W., Zhang, C., Cai, Z., Long, X., Chen, H., Wang, K., Yu, G., Shen, C., Shen, S.: Metric3d v2: A versatile monocular geometric foundation model for zero-shot metric depth and surface normal estimation. arXiv preprint arXiv:2404.15506  (2024)

\bibitem{jain2021putting_nerf_on_a_diet}
Jain, A., Tancik, M., Abbeel, P.: Putting nerf on a diet: Semantically consistent few-shot view synthesis. In: Proceedings of the IEEE/CVF International Conference on Computer Vision. pp. 5885--5894 (2021)

\bibitem{Kerbl20233DGS}
Kerbl, B., Kopanas, G., Leimkuehler, T., Drettakis, G.: 3d gaussian splatting for real-time radiance field rendering. ACM Transactions on Graphics (TOG)  \textbf{42},  1 -- 14 (2023)

\bibitem{knapitsch2017tanks}
Knapitsch, A., Park, J., Zhou, Q.Y., Koltun, V.: Tanks and temples: Benchmarking large-scale scene reconstruction. ACM Transactions on Graphics (ToG)  \textbf{36}(4),  1--13 (2017)

\bibitem{kuhn2020deepc}
Kuhn, A., Sormann, C., Rossi, M., Erdler, O., Fraundorfer, F.: Deepc-mvs: Deep confidence prediction for multi-view stereo reconstruction. In: 2020 International Conference on 3D Vision (3DV). pp. 404--413. Ieee (2020)

\bibitem{lee2021patchmatch}
Lee, J.Y., DeGol, J., Zou, C., Hoiem, D.: Patchmatch-rl: Deep mvs with pixelwise depth, normal, and visibility. In: Proceedings of the IEEE/CVF International Conference on Computer Vision. pp. 6158--6167 (2021)

\bibitem{lee2022qff}
Lee, J.Y., Wu, Y., Zou, C., Wang, S., Hoiem, D.: Qff: Quantized fourier features for neural field representations. arXiv preprint arXiv:2212.00914  (2022)

\bibitem{li2022nerfacc}
Li, R., Tancik, M., Kanazawa, A.: Nerfacc: A general nerf accleration toolbox. arXiv preprint arXiv:2210.04847  (2022)

\bibitem{li2023neuralangelo}
Li, Z., M\"uller, T., Evans, A., Taylor, R.H., Unberath, M., Liu, M.Y., Lin, C.H.: Neuralangelo: High-fidelity neural surface reconstruction. In: IEEE Conference on Computer Vision and Pattern Recognition ({CVPR}) (2023)

\bibitem{lorensen1987marching}
Lorensen, W.E., Cline, H.E.: Marching cubes: A high resolution 3d surface construction algorithm. SIGGRAPH '87, Association for Computing Machinery, New York, NY, USA (1987). \doi{10.1145/37401.37422}

\bibitem{lucas1981iterative}
Lucas, B.D., Kanade, T.: An iterative image registration technique with an application to stereo vision. In: IJCAI'81: 7th international joint conference on Artificial intelligence. vol.~2, pp. 674--679 (1981)

\bibitem{ma2021eppmvsnet}
Ma, X., Gong, Y., Wang, Q., Huang, J., Chen, L., Yu, F.: Epp-mvsnet: Epipolar-assembling based depth prediction for multi-view stereo. In: Proceedings of the IEEE/CVF International Conference on Computer Vision. pp. 5732--5740 (2021)

\bibitem{manivasagam2020lidarsim}
Manivasagam, S., Wang, S., Wong, K., Zeng, W., Sazanovich, M., Tan, S., Yang, B., Ma, W.C., Urtasun, R.: Lidarsim: Realistic lidar simulation by leveraging the real world. In: CVPR (2020)

\bibitem{mildenhall2020nerf}
Mildenhall, B., Srinivasan, P.P., Tancik, M., Barron, J.T., Ramamoorthi, R., Ng, R.: Nerf: Representing scenes as neural radiance fields for view synthesis. In: ECCV. Springer, Cham (2020)

\bibitem{mildenhall2021nerf}
Mildenhall, B., Srinivasan, P.P., Tancik, M., Barron, J.T., Ramamoorthi, R., Ng, R.: Nerf: Representing scenes as neural radiance fields for view synthesis. Communications of the ACM  \textbf{65}(1),  99--106 (2021)

\bibitem{muller2022instant}
M{\"u}ller, T., Evans, A., Schied, C., Keller, A.: Instant neural graphics primitives with a multiresolution hash encoding. arXiv  (2022)

\bibitem{niemeyer2022regnerf}
Niemeyer, M., Barron, J.T., Mildenhall, B., Sajjadi, M.S., Geiger, A., Radwan, N.: Regnerf: Regularizing neural radiance fields for view synthesis from sparse inputs. In: Proceedings of the IEEE/CVF Conference on Computer Vision and Pattern Recognition. pp. 5480--5490 (2022)

\bibitem{park2020non}
Park, J., Joo, K., Hu, Z., Liu, C.K., So~Kweon, I.: Non-local spatial propagation network for depth completion. In: Computer Vision--ECCV 2020: 16th European Conference, Glasgow, UK, August 23--28, 2020, Proceedings, Part XIII 16. pp. 120--136. Springer (2020)

\bibitem{radford_icml2021_clip}
Radford, A., Kim, J.W., Hallacy, C., Ramesh, A., Goh, G., Agarwal, S., Sastry, G., Askell, A., Mishkin, P., Clark, J., Krueger, G., Sutskever, I.: Learning transferable visual models from natural language supervision. In: ICML (2021)

\bibitem{Rho_2023_CVPR}
Rho, D., Lee, B., Nam, S., Lee, J.C., Ko, J.H., Park, E.: Masked wavelet representation for compact neural radiance fields. In: Proceedings of the IEEE/CVF Conference on Computer Vision and Pattern Recognition (CVPR). pp. 20680--20690 (June 2023)

\bibitem{roessle2022dense}
Roessle, B., Barron, J.T., Mildenhall, B., Srinivasan, P.P., Nie{\ss}ner, M.: Dense depth priors for neural radiance fields from sparse input views. In: Proceedings of the IEEE/CVF Conference on Computer Vision and Pattern Recognition. pp. 12892--12901 (2022)

\bibitem{colmapsfm}
Sch{\"o}nberger, J.L., Frahm, J.M.: Structure-from-motion revisited. In: Proceedings of the IEEE Conference on Computer Vision and Pattern Recognition (CVPR) (2016)

\bibitem{schoenberger2016mvs}
Sch\"{o}nberger, J.L., Zheng, E., Pollefeys, M., Frahm, J.M.: Pixelwise view selection for unstructured multi-view stereo. In: European Conference on Computer Vision (ECCV) (2016)

\bibitem{colmapmvs}
Sch{\"o}nberger, J.L., Zheng, E., Pollefeys, M., Frahm, J.M.: Pixelwise view selection for unstructured multi-view stereo. In: Proceedings of the European Conference on Computer Vision (ECCV) (2016)

\bibitem{schops2017multi}
Schops, T., Schonberger, J.L., Galliani, S., Sattler, T., Schindler, K., Pollefeys, M., Geiger, A.: A multi-view stereo benchmark with high-resolution images and multi-camera videos. In: Proceedings of the IEEE Conference on Computer Vision and Pattern Recognition. pp. 3260--3269 (2017)

\bibitem{robust_mvd}
Schr{\"{o}}ppel, P., Bechtold, J., Amiranashvili, A., Brox, T.: A benchmark and a baseline for robust multi-view depth estimation. In: Proceedings of the International Conference on 3D Vision (3DV). pp. 637--645 (2022)

\bibitem{deepv2d}
Teed, Z., Deng, J.: Deepv2d: Video to depth with differentiable structure from motion. In: International Conference on Learning Representations (ICLR) (2020)

\bibitem{truong2023sparf}
Truong, P., Rakotosaona, M.J., Manhardt, F., Tombari, F.: Sparf: Neural radiance fields from sparse and noisy poses. In: CVPR. pp. 4190--4200 (2022)

\bibitem{demon}
Ummenhofer, B., Zhou, H., Uhrig, J., Mayer, N., Ilg, E., Dosovitskiy, A., Brox, T.: {DeMoN}: Depth and motion network for learning monocular stereo. In: Proceedings of the IEEE Conference on Computer Vision and Pattern Recognition (CVPR). pp. 5622--5631 (2017)

\bibitem{verbin2022ref}
Verbin, D., Hedman, P., Mildenhall, B., Zickler, T., Barron, J.T., Srinivasan, P.P.: Ref-nerf: Structured view-dependent appearance for neural radiance fields. In: 2022 IEEE/CVF Conference on Computer Vision and Pattern Recognition (CVPR). pp. 5481--5490. IEEE (2022)

\bibitem{wang2023sparsenerf}
Wang, G., Chen, Z., Loy, C.C., Liu, Z.: Sparsenerf: Distilling depth ranking for few-shot novel view synthesis. In: Proceedings of the IEEE/CVF International Conference on Computer Vision. pp. 9065--9076 (2023)

\bibitem{wang2021neus}
Wang, P., Liu, L., Liu, Y., Theobalt, C., Komura, T., Wang, W.: Neus: Learning neural implicit surfaces by volume rendering for multi-view reconstruction. Advances in Neural Information Processing Systems (NeurIPS)  (2021)

\bibitem{Wang2023DUSt3RG3}
Wang, S., Leroy, V., Cabon, Y., Chidlovskii, B., Revaud, J.: Dust3r: Geometric 3d vision made easy. In: Proceedings of the IEEE Conference on Computer Vision and Pattern Recognition (CVPR) (2024)

\bibitem{wei2021nerfingmvs}
Wei, Y., Liu, S., Rao, Y., Zhao, W., Lu, J., Zhou, J.: Nerfingmvs: Guided optimization of neural radiance fields for indoor multi-view stereo. In: Proceedings of the IEEE/CVF International Conference on Computer Vision. pp. 5610--5619 (2021)

\bibitem{wu2022diver}
Wu, L., Lee, J.Y., Bhattad, A., Wang, Y.X., Forsyth, D.: Diver: Real-time and accurate neural radiance fields with deterministic integration for volume rendering. In: CVPR (2022)

\bibitem{Xu2022Multi}
Xu, Q., Kong, W., Tao, W., Pollefeys, M.: Multi-scale geometric consistency guided and planar prior assisted multi-view stereo. IEEE Transactions on Pattern Analysis and Machine Intelligence  (2022)

\bibitem{yang2023freenerf}
Yang, J., Pavone, M., Wang, Y.: Freenerf: Improving few-shot neural rendering with free frequency regularization. In: Proceedings of the IEEE/CVF Conference on Computer Vision and Pattern Recognition. pp. 8254--8263 (2023)

\bibitem{MVS2D}
Yang, Z., Ren, Z., Shan, Q., Huang, Q.: Mvs2d: Efficient multiview stereo via attention-driven 2d convolutions. In: Proceedings of the IEEE Conference on Computer Vision and Pattern Recognition (CVPR). pp. 8564--8574 (2022)

\bibitem{yao2018mvsnet}
Yao, Y., Luo, Z., Li, S., Fang, T., Quan, L.: Mvsnet: Depth inference for unstructured multi-view stereo. In: Proceedings of the European conference on computer vision (ECCV). pp. 767--783 (2018)

\bibitem{mvsnet}
Yao, Y., Luo, Z., Li, S., Fang, T., Quan, L.: Mvsnet: Depth inference for unstructured multi-view stereo. In: Proceedings of the European Conference on Computer Vision (ECCV) (2018)

\bibitem{yao2019recurrent}
Yao, Y., Luo, Z., Li, S., Shen, T., Fang, T., Quan, L.: Recurrent mvsnet for high-resolution multi-view stereo depth inference. In: Proceedings of the IEEE/CVF conference on computer vision and pattern recognition. pp. 5525--5534 (2019)

\bibitem{Yariv2021VolumeRO}
Yariv, L., Gu, J., Kasten, Y., Lipman, Y.: Volume rendering of neural implicit surfaces. ArXiv  \textbf{abs/2106.12052} (2021)

\bibitem{yu2021pixelnerf}
Yu, A., Ye, V., Tancik, M., Kanazawa, A.: pixelnerf: Neural radiance fields from one or few images. In: CVPR (2021)

\bibitem{yu2022monosdf}
Yu, Z., Peng, S., Niemeyer, M., Sattler, T., Geiger, A.: Monosdf: Exploring monocular geometric cues for neural implicit surface reconstruction. Advances in neural information processing systems  \textbf{35},  25018--25032 (2022)

\bibitem{vis-mvsnet}
Zhang, J., Li, S., Luo, Z., Fang, T., Yao, Y.: Vis-mvsnet: Visibility-aware multi-view stereo network. International Journal of Computer Vision (IJCV)  \textbf{131}(1),  199--214 (2023)

\bibitem{zhang2018perceptual}
Zhang, R., Isola, P., Efros, A.A., Shechtman, E., Wang, O.: The unreasonable effectiveness of deep features as a perceptual metric. In: CVPR (2018)

\end{thebibliography}
}
\onecolumn
\clearpage
\setcounter{page}{1}
\begin{center}
{\Large{Supplementary Material}}
\end{center}

\section{Training Details}
We provide the detailed network architecture in Table~\ref{tab:network}. 
Our model contains a total of 16,385,392 parameters and is trained for 50,000 steps on ETH3D in 2.25 hours and 200,000 steps on TanksandTemples in 9 hours using a single Nvidia A40 GPU. 

\begin{table}[h]
\centering
\caption{
    \textbf{Network Architecture Details.} We have a spatial feature extractor that uses implementation of QFF~\cite{lee2022qff} to extract features for each point. The extracted features are passed into the Density MLP to obtain per-point density and geometric features of length 15. The extracted geometry features are passed into Color MLP (along with the direction) and Surface Normal MLP to extract color $\mathbf{c}$ and surface normal $\mathbf{n}_\theta$ respectively. 
}
\begin{small}
\resizebox{0.5\textwidth}{!}{
    \begin{tabular}{@{}c|c|c|c@{}}
    \toprule
    Name &  \# Parameters & Input & Output Size \\
    \midrule
    \multicolumn{4}{c}{\textbf{Spatial Features}} \\
    \midrule
    QFF~\cite{lee2022qff} & 32x80x80x80     & $(x, y, z) \in \mathbb{R}^3$ & 32 \\
    \midrule
    \multicolumn{4}{c}{\textbf{Density MLP}} \\
    \midrule
    $D0_{\theta}$ & 32x16 & QFF & 16 \\
    $D1_{\theta}$ & 16x15 & $ReLU(D0_\theta)$ & 15 \\
    $D2_{\theta}$ & 16x1 & $ReLU(D0_\theta)$ & $\sigma \in \mathbb{R}^1$ \\
    \midrule
    \multicolumn{4}{c}{\textbf{Color MLP}} \\
    \midrule
    $C0_{\theta}$ & 18x16 & $D1_{\theta} + (\mathbf{d} \in \mathbb{R}^3$) & 16 \\
    $C1_{\theta}$ & 16x3 & $ReLU(C0_\theta)$ & $\mathbf{c} \in \mathbb{R}^3$  \\
    \midrule
    \multicolumn{4}{c}{\textbf{Surface Normal MLP}} \\
    \midrule
    $S0_{\theta}$ & 15x16 & $D1_{\theta} $ & 16 \\
    $S1_{\theta}$ & 16x3 & $ReLU(S0_\theta)$ & $\mathbf{n}_\theta \in \mathbb{R}^3$\\
    \bottomrule
    \end{tabular}
}
\label{tab:network}
\end{small}
\end{table}
\section{Baseline Training Details}
\begin{figure}[h!]
\centering
\begin{tabular}{c}
        {\includegraphics[width=0.95\textwidth]{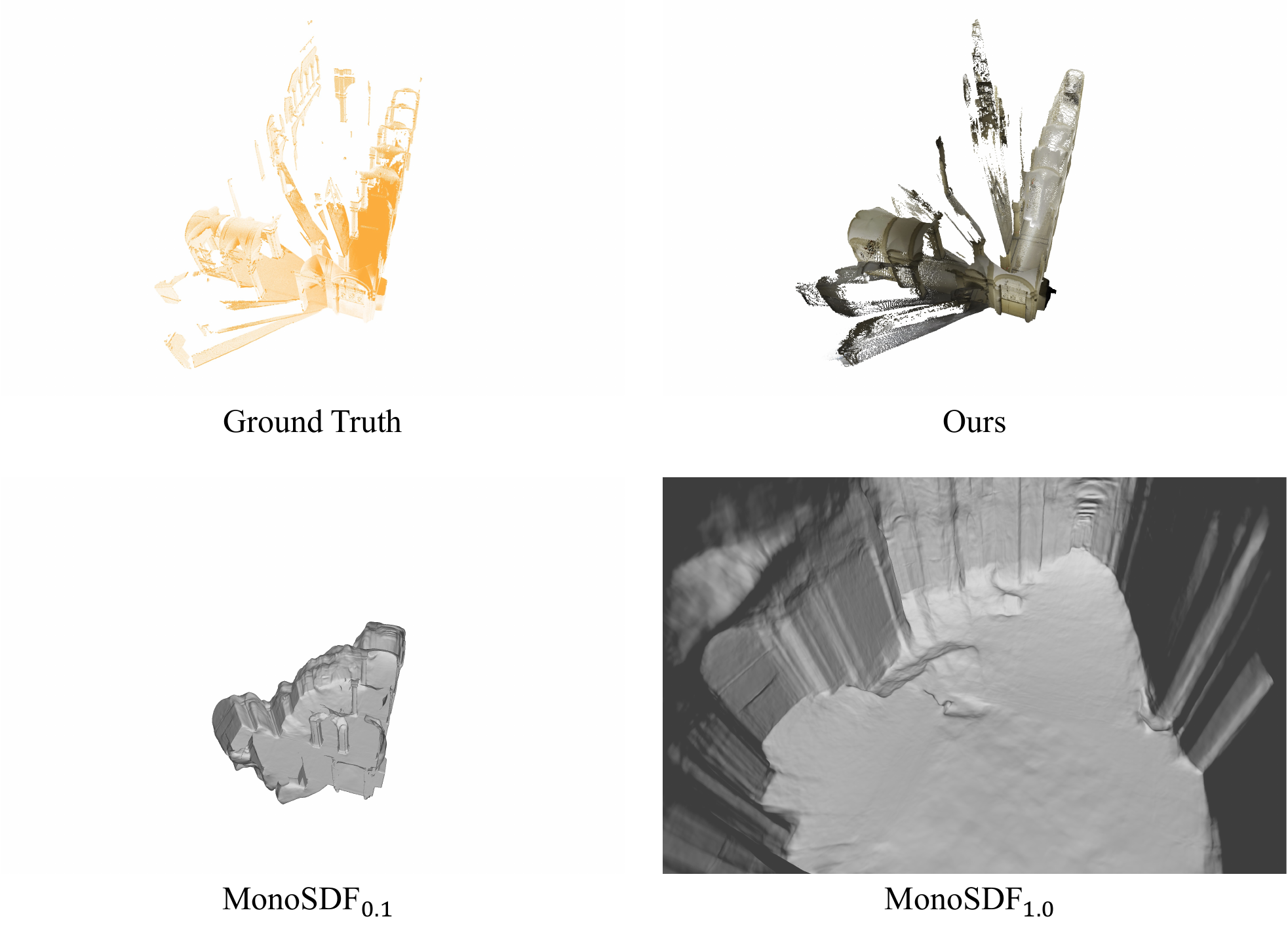}}
        
\end{tabular}
\vspace{-0.05in}
\caption{\textbf{Visualization of MonoSDF with different paramters in \textit{Relief\_2}.} 
    MonoSDF$_{1.0}$ denotes MonoSDF trained with the default bias parameter~(1.0) in the code provided by the authors on the large scale TnT~\cite{knapitsch2017tanks} evaluation. MonoSDF$_{0.1}$ denotes MonoSDF trained with the parameter suggested by the authors~(0.1) for large-scale scenes specifically.
    MonoSDF~\cite{yu2022monosdf} reconstructs better mesh given smaller bias, and falls into local minimum with original bias. 
}
\label{fig:monosdf_local_minimum}
\end{figure}

\noindent \textbf{MonoSDF}:
We show how varying the initialization bias of MonoSDF~\cite{yu2022monosdf} affects its reconstruction quality. 
We used the author provided configs of TnT on ETH3D, but found that MonoSDF suffer from local minimum in some challenging scenes with original bias parameters due to the scene scale. We therefore contacted the MonoSDF authors and were advised to use a small bias for initialization.
Figure~\ref{fig:monosdf_local_minimum} compares the reconstruction of MonoSDF given different bias parameters in a challenging scene \textit{relief\_2} of ETH3D~\cite{schops2017multi}. 

\begin{figure}[h!]

\centering
\begin{tabular}{c}
\includegraphics[width=.99\textwidth]{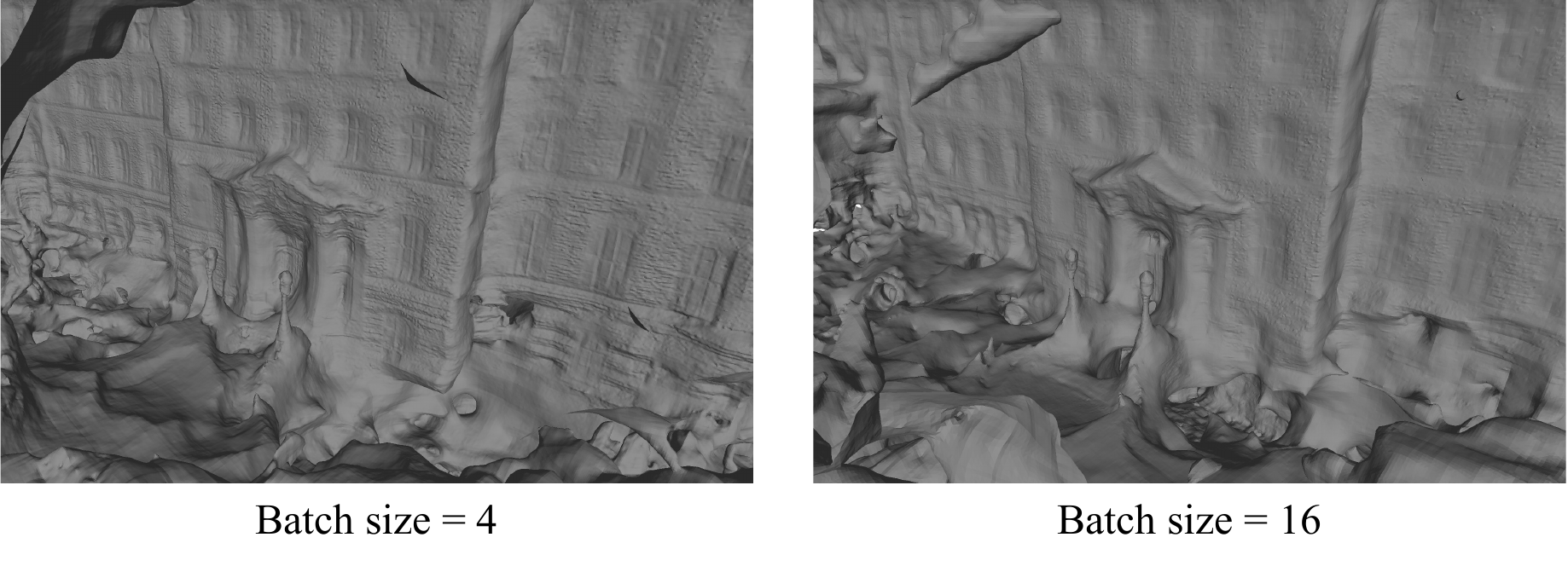}
        
\end{tabular}
\vspace{-0.05in}
\caption{\textbf{Visualization of Neuralangelo with different batch size in \textit{Facade}.}  
}
\label{fig:neuralangelo_comp}
\end{figure}

\noindent \textbf{Neuralangelo}:
For Neuralangelo~\cite{li2023neuralangelo} experiment on ETH3D~\cite{schops2017multi}, we follow author provided setup on TanksAndTemples~\cite{knapitsch2017tanks}, but use a batch size of 4 instead of 16 to run on the same device settings. 
We additionally disable image embedding features, as we empirically found it to worsen the results. 
One visualization of results with different batch size is present in Figure~\ref{fig:neuralangelo_comp}.  
The $F$-score$_{2cm}$, $F$-score$_{5cm}$ of high and low batch size results are 1.53, 11.5 and 1.46, 11.8.
        

\clearpage
\section{Depth Evaluation}
We provide evaluation of depth maps from radiance based methods and MVS methods in Tab.~\ref{tab:mvd} following robust multi-view benchmark~\cite{robust_mvd}. We report the Absolute Relative Error~(rel) and Inlier Ratio~($\tau$) with a threshold of 1.03, and split the evaluated methods based on the requirement for poses, depth ranges, and intrinsics. To better compare the performance, we additionally group NeRF models together. We show that though not as accurate as MVS methods, we predict more complete depths, as our rel is lower than all classical MVS methods. 
\begin{table*}[h]
\normalsize
\caption{
\textbf{Multi-view depth evaluation} with different settings: 
a) Classical approaches; 
b) with poses and depth range, without alignment;
c) absolute scale evaluation with poses, without depth range and alignment; 
d) without poses and depth range, but with alignment;
e) neural radiance field based models.
`med` means alignment based on median ground truth depth and the median predicted depth. 
Numbers with $^\ast$ are from our results, while others are from DUSt3R~\cite{Wang2023DUSt3RG3}.
The best results for each setting are in \textbf{bold}. 
}
\begin{center}
\renewcommand\arraystretch{1.2}
\setlength{\tabcolsep}{1pt} 
\small
\hspace{-3mm}
\resizebox{0.5\textwidth}{!}{
\begin{tabular}{llccccrr}
 \hline
\specialrule{1.5pt}{0.5pt}{0.5pt}
\multicolumn{2}{l}{\multirow{2}{*}{Methods}} & GT & GT & GT & Align & \multicolumn{2}{c}{ETH3D} \\
\cline{7-8}
 && Pose & Range & Intrinsics &  & rel $\downarrow$ & $\tau \uparrow$ \\
\specialrule{1.5pt}{0.5pt}{0.5pt}
\multirow{3}{*}{(a)} & COLMAP~\cite{colmapsfm,colmapmvs} & $\checkmark$ & $\times$ & $\checkmark$ & $\times$ & 16.4&55.1 \\
&COLMAP Dense~\cite{colmapsfm,colmapmvs} & $\checkmark$&$\times$ & $\checkmark$ & $\times$ & 89.8 & 23.2 \\
&ACMMP$^\ast$~\cite{Xu2022Multi} & $\checkmark$&$\times$ & $\checkmark$ & $\times$ & {\bf 16.0} & {\bf 91.6} \\
\hline
\multirow{5}{*}{(b)} & MVSNet~\cite{mvsnet} & $\checkmark$ & $\checkmark$ &$\checkmark$ & $\times$ & 35.4 & 31.4 \\
& MVSNet Inv. Depth~\cite{mvsnet} & $\checkmark$ & $\checkmark$ &$\checkmark$ & $\times$ & 21.6 & 35.6\\
& Vis-MVSSNet~\cite{vis-mvsnet} & $\checkmark$ & $\checkmark$ & $\checkmark$ & $\times$ &{\bf 10.8}&{\bf 43.3}\\
& MVS2D ScanNet~\cite{MVS2D} & $\checkmark$ & $\checkmark$ & $\checkmark$ & $\times$ & 27.4 & 4.8 \\
& MVS2D DTU~\cite{MVS2D} & $\checkmark$ & $\checkmark$& $\checkmark$ & $\times$ & 99.0 & 11.6 \\
\hline
\multirow{8}{*}{(c)} & DeMon~\cite{demon} & $\checkmark$ & $\times$ &$\checkmark$ & $\times$ & 19.0 & 16.2\\
& DeepV2D KITTI~\cite{deepv2d} & $\checkmark$ & $\times$ &$\checkmark$ & $\times$ & 30.1 & 9.4 \\
& DeepV2D ScanNet~\cite{deepv2d} & $\checkmark$ & $\times$ &$\checkmark$ & $\times$ & 18.7 & 28.7 \\
& MVSNet~\cite{mvsnet} & $\checkmark$ & $\times$ &$\checkmark$ & $\times$ & 507.7 & 8.3 \\
& MVSNet Inv. Depth~\cite{mvsnet} & $\checkmark$ & $\times$ &$\checkmark$ & $\times$ & 60.3 & 5.8 \\
& Vis-MVSNet \cite{vis-mvsnet} & $\checkmark$ & $\times$ &$\checkmark$ & $\times$ & 51.5 & 17.4 \\
& MVS2D ScanNet~\cite{MVS2D} & $\checkmark$ &$\times$ &$\checkmark$ &$\times$ & 30.7 & 14.4 \\
& MVS2D DTU~\cite{MVS2D} &$\checkmark$ & $\times$ & $\checkmark$ & $\times$ & 78.0 & 0.0 \\
& Robust MVD Baseline~\cite{robust_mvd} &$\checkmark$&$\times$&$\checkmark$&$\times$ & {\bf 9.0}&{\bf 42.6}\\
\hline
\multirow{6}{*}{(d)} & DeMoN~\cite{demon} &$\times$&$\times$ &$\checkmark$& $\|\mathbf{t}\|$ & 17.4 & 15.4 \\
& DeepV2D KITTI~\cite{deepv2d} &$\times$&$\times$&$\checkmark$&med&27.1 & 10.1 \\
& DeepV2D ScanNet~\cite{deepv2d} &$\times$&$\times$&$\checkmark$& med & 11.8 & 29.3 \\
& \bf{{\textbf{DUSt3R}} 224-NoCroCo\cite{Wang2023DUSt3RG3}} &$\times$&$\times$&$\times$&med& 9.51&40.07 \\
& \bf{{\textbf{DUSt3R}} 224\cite{Wang2023DUSt3RG3}}     &$\times$&$\times$&$\times$&med&4.71&61.74 \\
& \bf{{\textbf{DUSt3R}} 512\cite{Wang2023DUSt3RG3}} &$\times$&$\times$&$\times$&med&{\bf 2.91}&{\bf 76.91}\\
\hline
\multirow{4}{*}{(e)} & RegNeRF$^\ast$~\cite{niemeyer2022regnerf} &$\checkmark$&$\times$ &$\checkmark$& $\times$& 24.9 & 15.0 \\
& FreeNeRF$^\ast$~\cite{yang2023freenerf} &$\checkmark$&$\times$&$\checkmark$&$\times$&194.0 & 7.3 \\
& Ours$^\ast$ &$\checkmark$&$\times$&$\checkmark$& $\times$ & 7.4 & 67.2 \\
& Ours~(MVS-Depth)$^\ast$ &$\checkmark$&$\times$&$\checkmark$& $\times$ & {\bf 7.2} & {\bf 83.5} \\
\specialrule{1.5pt}{0.5pt}{0.5pt}
\end{tabular}}

\label{tab:mvd}
\end{center}
\vspace*{-1mm}
\end{table*}

\clearpage

\section{Sparse-view Tanks and Temples Comparison}
\begin{figure}[h]
\centering
\begin{tabular}{c}
        {\includegraphics[width=0.8\textwidth]{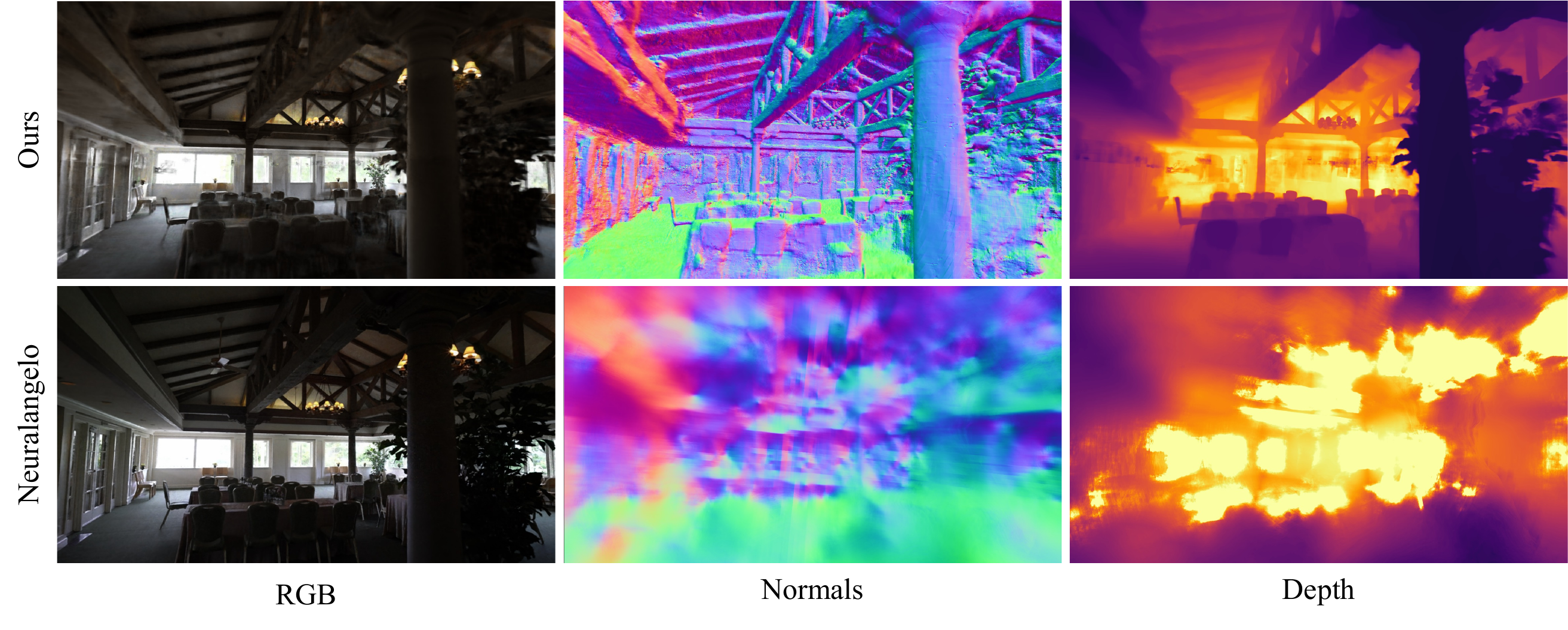}}
\end{tabular}
\vspace{-5mm}

\caption{\textbf{Training view visualization on sparse view TnT.} We compare the rendered color images, depth map, and normals map from the training view on \textit{Meetingroom} for our method and Neuralangelo~\cite{li2023neuralangelo}.}
\label{fig:tnt_sparse}
\end{figure}

We compare our method with Neuralangelo~\cite{li2023neuralangelo} on Tanks and Temples~\cite{knapitsch2017tanks} with 1/5 views that are uniformly sampled from original views in Tab.~\ref{tab:tnt_sparse}. We follow Neuralangelo to preprocess the scene, and use the same parameters~(include the batch size) as the original paper except for using 200,000 instead of 500,000 training steps due to reduced input images. Our method outperform Neuralangelo in all scenes, showing that our method works better for the challenging sparse view setup. See Fig~\ref{fig:tnt_sparse} for qualitative comparison.

\begin{table}[h]
\caption{
    \textbf{Comparison on TnT~\cite{knapitsch2017tanks} with sparse input views}. 
    We report the $F$-score of our method and Neuralangelo~\cite{li2023neuralangelo} in three large-scale TnT scenes following Neuralangelo preprocessing. Best results are in \textbf{bold}.
}
\centering
\vspace{-0.25em}
\resizebox{0.49\textwidth}{!}{

\begin{tabular}{c|ccc}
\toprule
 & Meetingroom & Courthouse & Barn \\
 \midrule
 Ours & {\bf 14.0} & {\bf 16.0} & {\bf 30.8} \\
 Neuralangelo & 1.7 & 7.7 & 5.7 \\
\bottomrule
\end{tabular}
}
\label{tab:tnt_sparse}
\vspace{-0.5em}
\end{table}

\section{Foreground Fattening}
\begin{figure}[b]
\centering
\begin{tabular}{c}
    \includegraphics[width=0.90\textwidth]{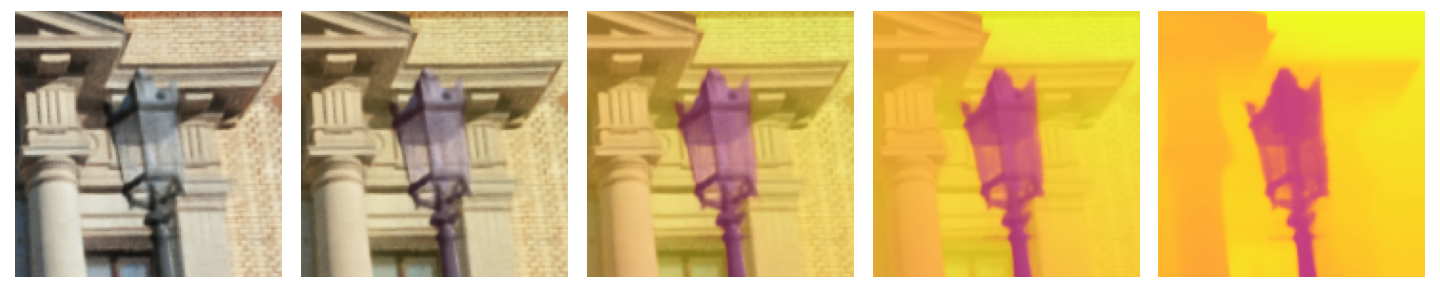}
\end{tabular}
\caption{
    \textbf{Image and Depth overlay visualization.} 
    From the left to right, we overlay the RGB image and the rendered depth map with varying fade thresholds. We show that our method does not experience foreground fattening as the images and the depths are precisely overlapped.
}
\label{fig:overlay}
\end{figure}

In patch-based MVS, Foreground fattening can happen due to the plane-based propagation of depth candidates and a patch-wise planar assumption in computing photometric scores. Our method does not suffer from foreground fattening because we do not make planar assumptions (current depth estimates are projected into other views to compute photometric scores), and the rendering loss discourages such artifacts. Figure~\ref{fig:overlay} shows the alignment of RGB image and the depth images. The image and the depth are aligned precisely, indicating that our method does not suffer from the foreground fattening.

\section{Additional Qualitative Results}
\begin{figure*}[h]
\centering
\begin{tabular}{c}
        \includegraphics[width=0.95\textwidth]{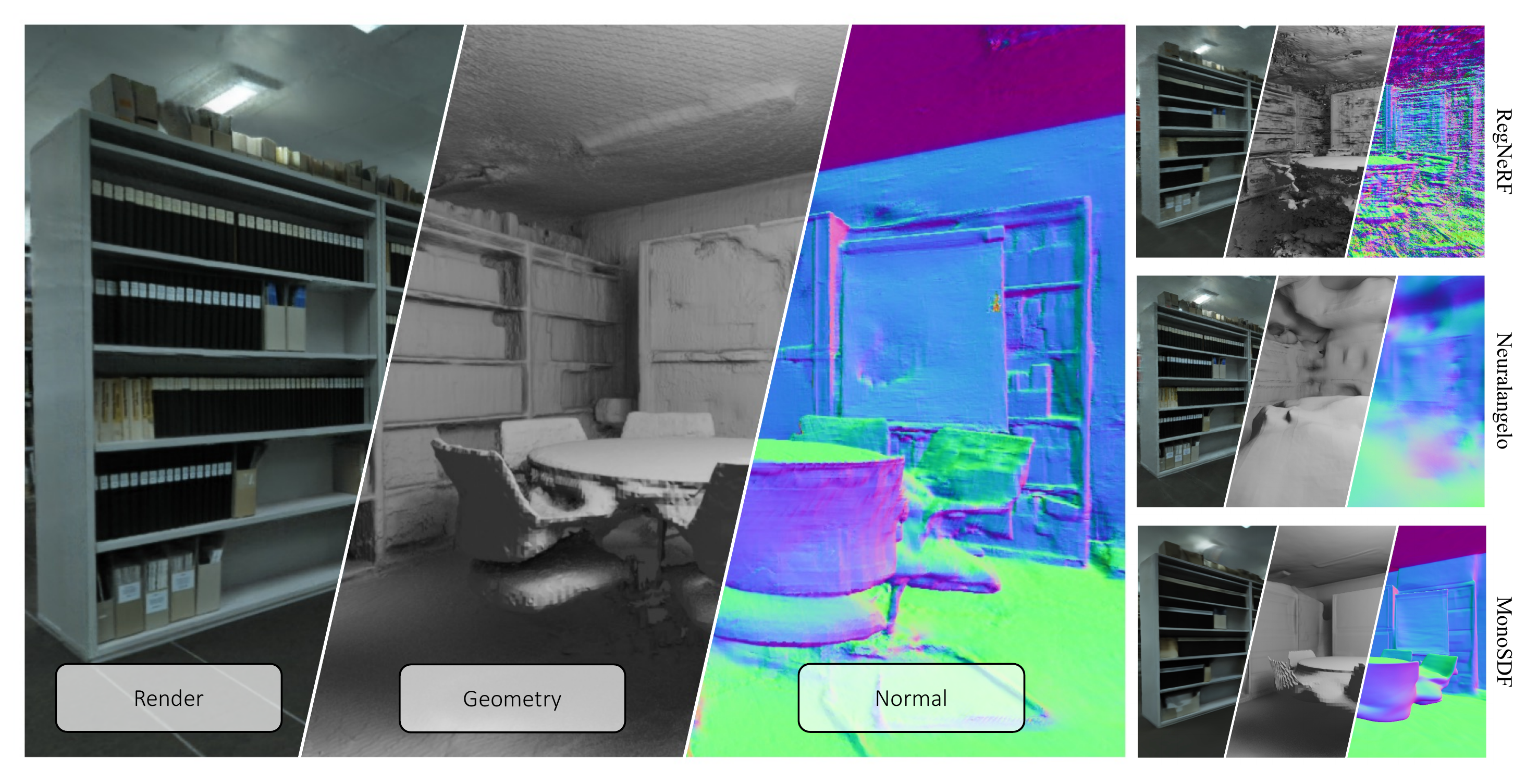}
\end{tabular}
\vspace{-0.05in}
\caption{\textbf{Additional qualitative comparison on \textit{kicker}.} We provide additional test view comparisons with baselines~\cite{niemeyer2022regnerf, yang2023freenerf, yu2022monosdf, li2023neuralangelo}.}
\label{fig:eth3d_additional_teaser}
\end{figure*}

\begin{figure}[h!]
\centering
\begin{tabular}{c}
        \includegraphics[width=0.95\textwidth]{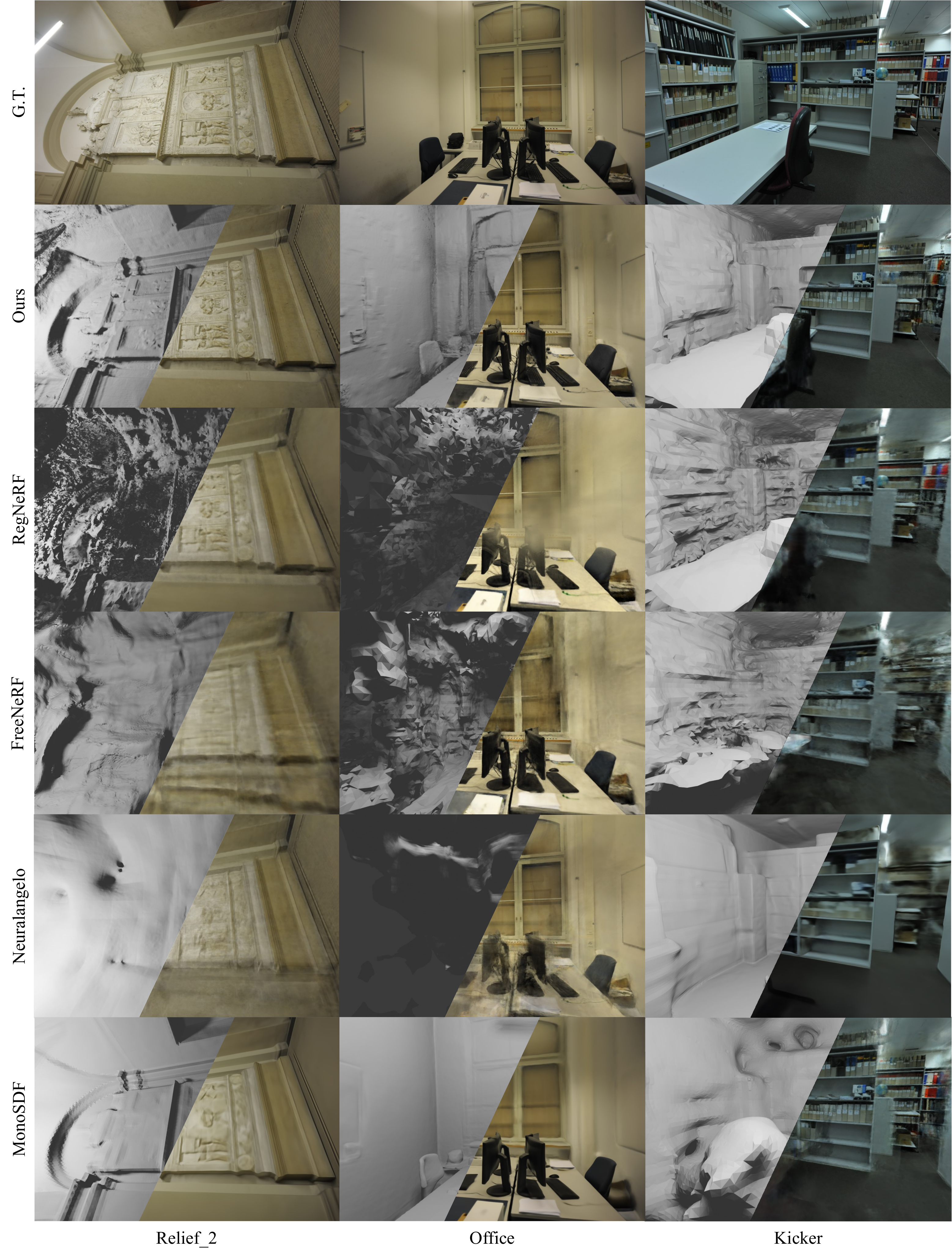}
        
\end{tabular}
\vspace{-0.05in}
    \caption{\textbf{Additional comparison of novel view images and meshes on ETH3D.} We provide additional comparisons with baselines~\cite{niemeyer2022regnerf, yang2023freenerf, yu2022monosdf, li2023neuralangelo}.
}
\label{fig:eth3d_additional_nvs}
\end{figure}

\begin{figure}[h!]
\centering
\begin{tabular}{c}
        \includegraphics[width=0.95\textwidth]{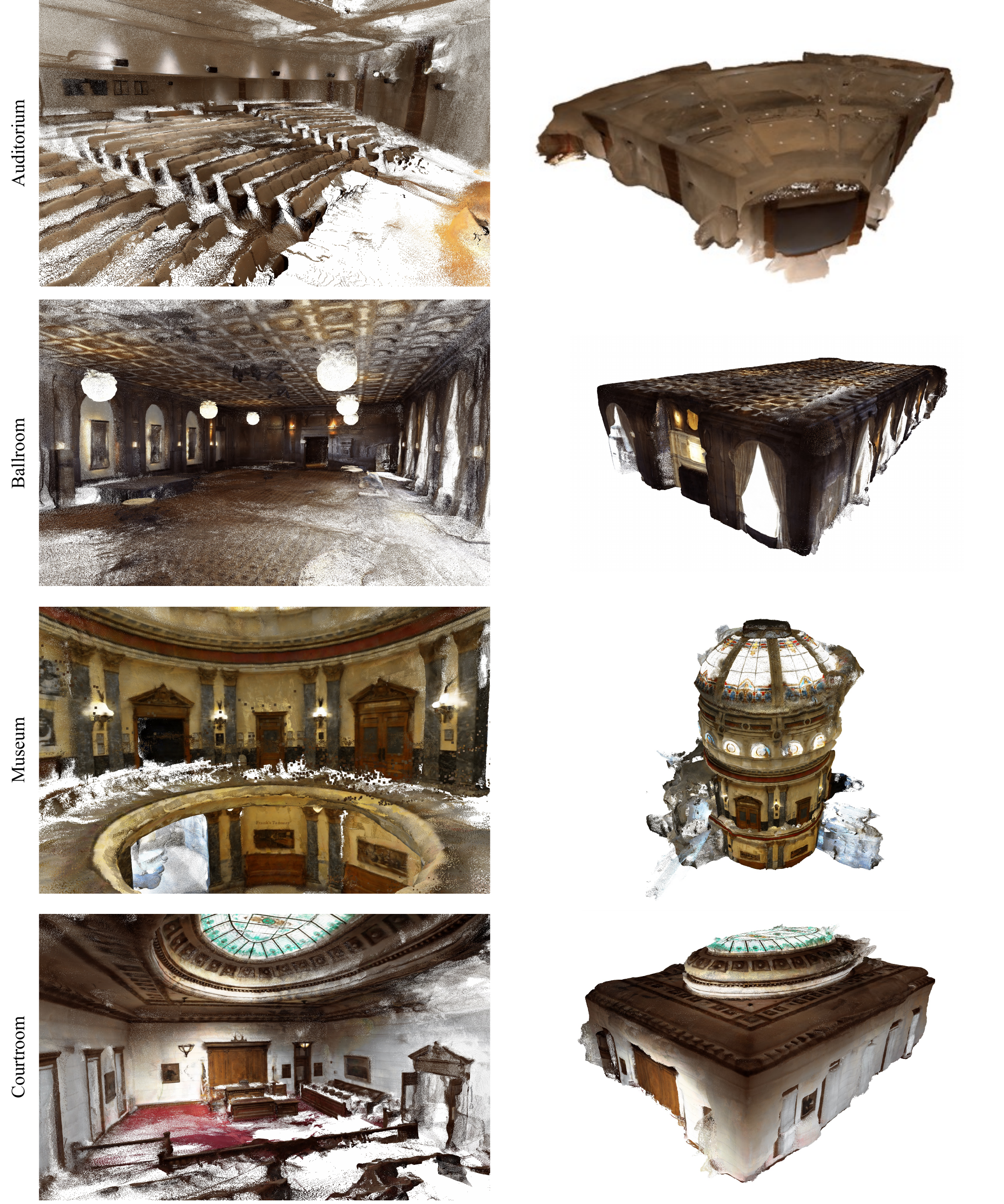}
        
\end{tabular}
\vspace{-0.05in}
    \caption{\textbf{Point clouds visualization for TnT advanced scenes~\cite{knapitsch2017tanks}.} 
    We visualize interior and far-away views for point clouds to have a better visualization of the reconstructed geometry. Our method reconstructs complete and accurate point clouds.
}
\label{fig:tnt_additional_pc}
\end{figure}

\begin{figure*}
\centering
\begin{tabular}{c}
        {\includegraphics[width=0.95\textwidth]{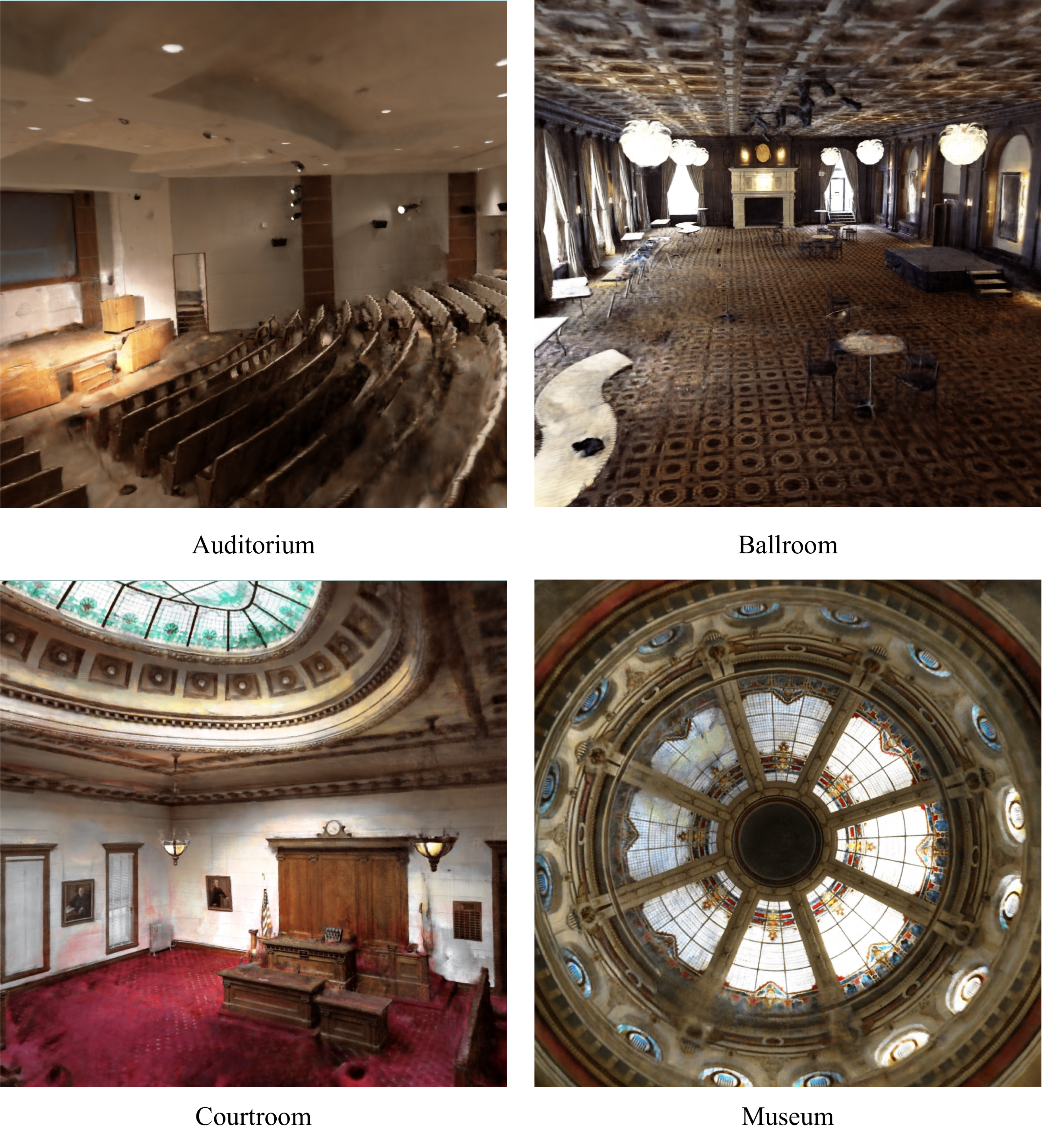}}
        
\end{tabular}
\vspace{-0.05in}
    \caption{\textbf{Free-form views rendering for TnT advanced scenes~\cite{knapitsch2017tanks}.} 
}
\label{fig:tnt_additional}
\end{figure*}

\begin{figure}[h!]
\centering
\begin{tabular}{c}
        \includegraphics[width=0.95\textwidth]{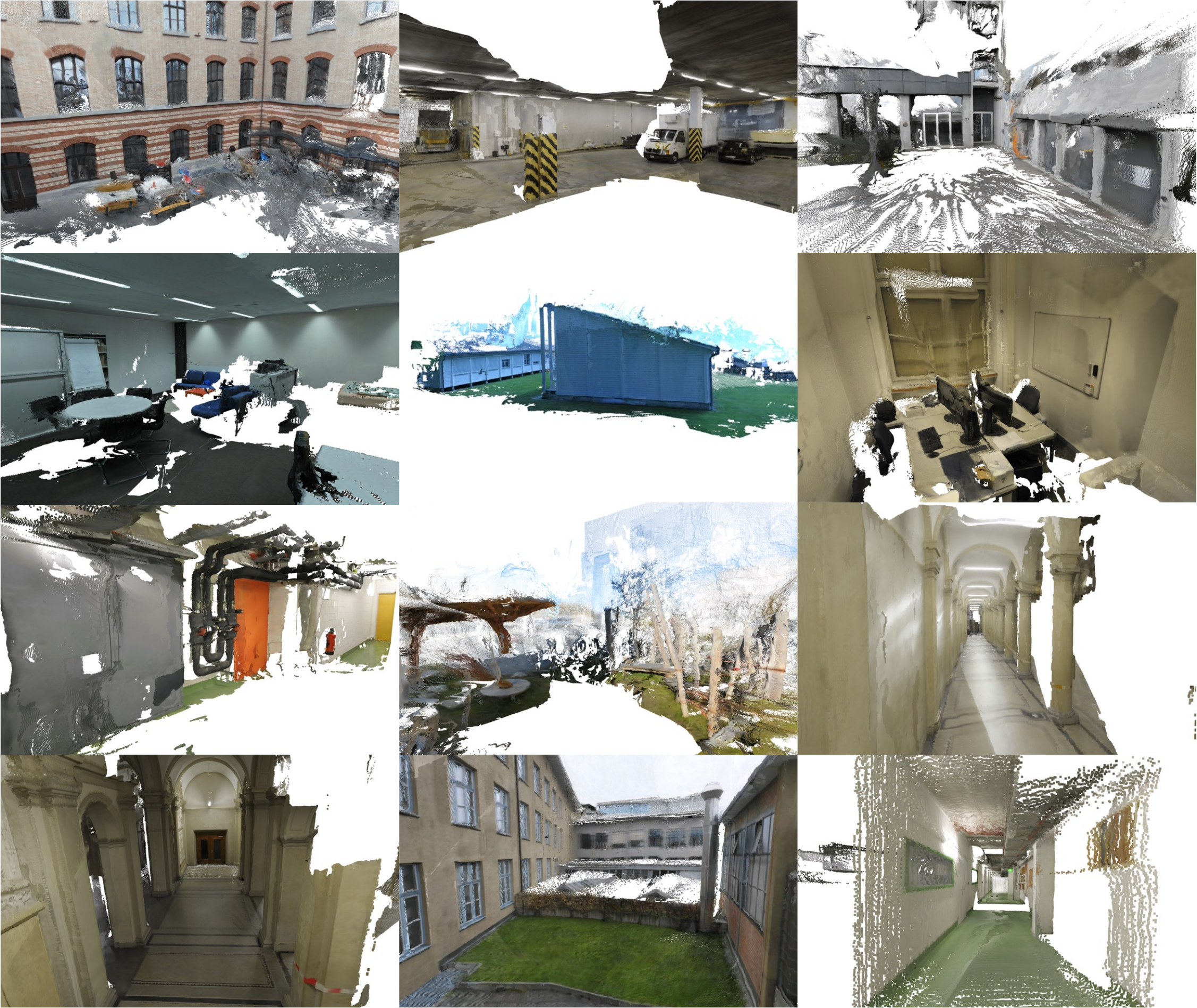}
        
\end{tabular}
\vspace{-0.05in}
    \caption{\textbf{Point clouds visualization for ETH3D.} 
    We visualize point clouds for all scenes on ETH3D~\cite{schops2017multi} except~\textit{Facade}. Our method reconstructs complete and accurate point clouds.
}
\label{fig:eth3d_ours_pc}
\end{figure}

\begin{figure}
\centering
\includegraphics[width=0.98\textwidth]{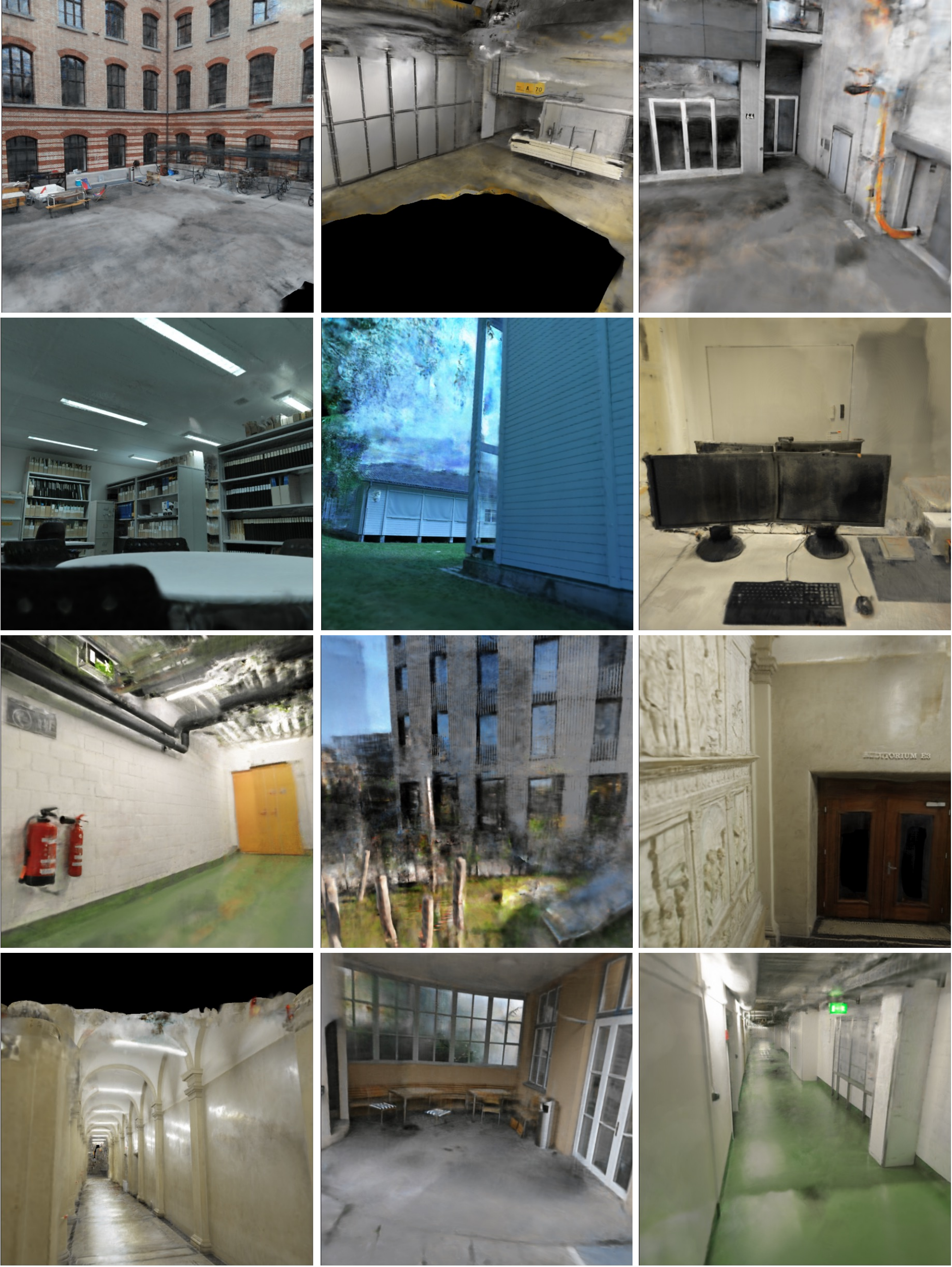}
\caption{\textbf{Free-form views rendering for ETH3D~\cite{schops2017multi}.} 
    We render novel views using our free-form viewer for each scenes in ETH3D except \textit{Facade}. 
    We show that our novel view rendering retains high-quality detailed textures in novel views. 
}
\label{fig:eth3d_additional}
\end{figure}

\noindent \textbf{Images}: 
We present one additional compairson like our teaser figure in Figure~\ref{fig:eth3d_additional_teaser}. Additional visualization of mesh and novel view synthesis are shown in Figure~\ref{fig:eth3d_additional_nvs}. 
We also provide additional visualizations of our method on subsets of TanksAndTemples~\cite{knapitsch2017tanks} advanced scenes in Figure~\ref{fig:tnt_additional_pc} and Figure~\ref{fig:tnt_additional}, and on ETH3D~\cite{schops2017multi} in Figure~\ref{fig:eth3d_ours_pc} and Figure~\ref{fig:eth3d_additional}.

\noindent \textbf{Videos}: 
We present a free-view rendering of scenes \textit{Relief\_2, Facade,} and \textit{Kicker} on ETH3D~\cite{schops2017multi} with trajectory interpolated from training poses.
We also provide comparisons with RegNeRF~\cite{niemeyer2022regnerf}, Neuralangelo\cite{li2023neuralangelo}, and MonoSDF~\cite{yu2022monosdf}. 
There are only 31, 76, 31 training views for the scenes, but the rendering is realistic and the video is smooth.
All our results, except for ones annotated with~(Ours-MVS-Depth) in Figure~{7} and Table~{1} of the main paper, are from our model with monocular cues.

\end{document}